\documentclass{bmvc2k}

\title{Biologically Plausible Variational Policy Gradient with Spiking Recurrent Winner-Take-All Networks}

\addauthor{Zhile Yang}{sczy@leeds.ac.uk}{1}
\addauthor{Shangqi Guo$^\dagger$}{shangqi_guo@foxmail.com}{2}
\addauthor{Ying Fang}{fy20@fjnu.edu.cn}{3}
\addauthor{Jian K. Liu$^\dagger$}{ j.liu9@leeds.ac.uk}{1}

\addinstitution{
    School of Computing,\\
    University of Leeds,\\
    Leeds, UK
}
\addinstitution{
    Department of Precision Instrument and Department of Automation,\\
    Tsinghua University,\\
    Beijing, China
}
\addinstitution{
    College of Computer and Cyber Security,\\
    Fujian Normal University,\\
    Fujian, China
}

\runninghead{Yang, Guo, Fang, Liu}{Biologically Plausible VPG with Spiking RWTA Nets}

\usepackage{algorithm}
\usepackage{algorithmic}
\usepackage{amsmath}
\usepackage{amssymb}
\usepackage{amsthm}
\usepackage{graphicx}
\usepackage{newtxtext,newtxmath}

\newtheorem{theorem}{Theorem}

\usepackage{wrapfig}
\usepackage{booktabs}
\usepackage{color}

\newcommand\blfootnote[1]{%
	\begingroup
	\renewcommand\thefootnote{}\footnote{#1}%
	\addtocounter{footnote}{-1}%
	\endgroup
}

\begin{document}

\maketitle
\blfootnote{$^\dagger$ Correspondence}

\begin{abstract}
	One stream of reinforcement learning research is exploring biologically plausible models and algorithms to simulate biological intelligence and fit neuromorphic hardware. Among them, reward-modulated spike-timing-dependent plasticity (R-STDP) is a recent branch with good potential in energy efficiency. However, current R-STDP methods rely on heuristic designs of local learning rules, thus requiring task-specific expert knowledge. In this paper, we consider a spiking recurrent winner-take-all network, and propose a new R-STDP method, spiking variational policy gradient (SVPG), whose local learning rules are derived from the global policy gradient and thus eliminate the need for heuristic designs. In experiments of MNIST classification and Gym InvertedPendulum, our SVPG achieves good training performance, and also presents better robustness to various kinds of noises than conventional methods. Code can be found at \url{https://github.com/yzlc080733/BMVC2022_SVPG}.

\end{abstract}

\section{Introduction}
Reinforcement learning (RL) with artificial neural networks (ANNs) has been applied to various scenarios \cite{devo_towards_2020,bellemare_autonomous_2020}. However, the adaptability, robustness, learning speed, etc. are still not satisfying. Recently some research build \textit{biologically plausible} learning systems similar to humans' brains to achieve human-level performances \cite{bing2019end,asgari2020low,wunderlich2019demonstrating}. In these brain-inspired methods, neural spiking signals are the key elements, which have demonstrated a great advantage for visual coding~\cite{yu_toward_2020, Gallego2022}. The enriched analysis of neural signals using RL also provide insights for explaining the formation of capabilities of brains in the field of computational neuroscience~\cite{Suhaimi2022}.

One mainstream biologically plausible method is spiking neural networks (SNNs). Current SNN RL methods can be divided into three categories: ANN-to-SNN conversion ANN2SNN) \cite{patel2019improved}, heuristically approximated backpropagation \cite{bellec_solution_2020,yanguas-gil_coarse_2020}, and modulated STDP (e.g., R-STDP) \cite{fremaux_neuromodulated_2016,yuan2019reinforcement}. Compared with the first two, R-STDP has the advantages of being biologically plausible in both training and testing, and being energy efficient. Recent R-STDP methods consider simulating the activity patterns of biological neurons, which are characterized by types of interactions between neurons, strengths of modulation, and modulation timings \cite{fremaux_neuromodulated_2016}, and have outperformed other state-of-the-art algorithms in many tasks in terms of control capabilities and energy efficiency \cite{bing2019end,asgari2020low,wunderlich2019demonstrating}. However, one drawback of them is their reliance on heuristic designs of local learning rules, since there is a gap between the local learning rules and the overall RL optimization target. This makes it inconvenient to apply R-STDP to RL tasks.

To bridge the gap, we adopt the variational inference method to study spiking neural networks, and propose a novel method, named spiking variational policy gradient (SVPG). It has been shown that variational inference can decompose the global target of a pattern generation or classification task into local learning rules \cite{rezende_variational_2011,Guo2019, jang_vowel_2020}. Besides, there is evidence that variational inference can be made biologically plausible using specific network structures \cite{rezende_variational_2011,beck_complex_2012,jang_vowel_2020}. As far as we know, this paper is the first to investigate variational inference for R-STDP.

To establish the theoretical relationship between local R-STDP rules and the global target, we propose a spiking neural network recurrently constructed of winner-take-all (WTA) circuits \cite{yu_emergent_2018, Guo2019} (named \textit{RWTA network}), together with an energy-based policy formulation. Under these settings, we prove that the fixed point of the RWTA network equals the inference result of the RL policy distribution, and that the R-STDP mechanism is a first-order approximation of the policy gradient. These bridge the R-STDP learning rule with the overall target. We apply SVPG to two typical RL tasks, including reward-based MNIST classification and Gym InvertedPendulum. SVPG successfully solved the tasks, achieving similar performance as a conventional ANN-based method. We also tested the robustness of the trained policies against various kinds of input noise \cite{patel2019improved}, network parameter noise \cite{li_robustness_2020}, and environment variation \cite{app10249013}. Results show SVPG has better robustness to the tested noises than popular biologically plausible RL methods.

\section{Preliminaries and Notations}
We begin with notations and formulations of the policy gradient algorithm, the SNN model, and the R-STDP framework.

\subsection{Policy Gradient} \label{sec_RL_PG}
In RL, Markov decision processes (MDPs) are commonly used for task formulation \cite{sutton_reinforcement_2018}. Here we adopt the notations from \cite{sutton_reinforcement_2018} and use tuple $ \langle \mathcal{S}, \mathcal{A}, P, R, \gamma\rangle $ to denote an MDP. The training objective is to find a policy $ \pi:\mathcal{S}\times\mathcal{A}\rightarrow [0,1] $ that maximizes the expected return $ J=\mathbb{E}_{\tau \sim \pi}\sum_{t=0}^{T-1}\gamma^t r_t $, where $ \tau $ is the sampled trajectory (denoted as $ \langle s_0, a_0, r_0, s_1, a_1, \dots, r_{T-1}, s_{T}\rangle $) and $ T $ is the total length of an episode.

Policy gradient is a widely used branch of algorithms in RL and has many popular derivatives \cite{lillicrap_continuous_2016,mnih_asynchronous_2016,schulman_proximal_2017}. A fundamental form of policy gradient is the REINFORCE algorithm, which optimizes the expected return through ascending the following gradient \cite{sutton_reinforcement_2018}:
\begin{equation}\label{eq_reinforce_gradient}
	\nabla_\theta J(\pi_\theta)=\mathbb{E}_{\tau\sim\pi_{\theta}}[\textstyle\sum_{t=0}^{T-1}\gamma^{t} r_t \textstyle\sum_{k=0}^{T-1}\nabla_\theta \log \pi_\theta(a_k|s_k)],
\end{equation}
where $ \theta $ is the parameters of the policy $ \pi $. In this paper we propose a new model for $ \pi_\theta $, so the key point is to derive the gradient $ \nabla_\theta\log{\pi_\theta(a_k | s_k)} $.

\subsection{Spiking Neurons}
Various neuron models have been proposed in the field of SNN. Here we consider the leaky integrate-and-fire (LIF) spike response model \cite{gerstner_neuronal_2014}. We consider that all spikes are binary and happen at discrete time steps. At each spike time step $ l $, the firing probability $ \rho $ of a neuron is determined by its membrane potential $ u (l) $:
\begin{equation}\label{eq_firing}
	\rho(l) = \exp \{u(l)-u_{\rm th}\},
\end{equation}
where $ u_{\rm th} $ is a threshold voltage. Note that $l$ is different from the RL step $t$. We consider that each $t$-step corresponds to a certain number of spike time steps. The membrane potential $ u(l) $ is determined by the spike train $ S $ from the presynaptic neurons \cite{gerstner_neuronal_2014}:
\begin{equation}\label{eq_potential}
	u(l)=\textstyle\sum_{j\in N(\cdot)} w_j \textstyle\int_{0}^{\infty}\kappa(y)S_j(l-y){\rm d}y+b,
\end{equation}
where $ N(\cdot) $ denotes the set of presynaptic neurons of the considered neuron, $ w_j $ is the synapse weight, $ \kappa $ is the excitatory postsynaptic potential, $ S_j $ is the spike train from neuron $ j $, and $ b $ is the self-excitation voltage of the considered neuron.


R-STDP models the learning of synapse weights given an external reward and local firing activities. We use $ w_{ij} $ to denote the weight of synapse between presynaptic neuron $ i $ and postsynaptic neuron $ j $. The R-STDP learning takes the form of $ \Delta w_{ij}=R(l)\cdot {\rm STDP}(l)$, where $ R(l) $ is the external reward signal and $ {\rm STDP}(l) $ is a coefficient determined by the STDP learning rule in the following form:
\begin{equation}\label{eq_STDP_form}
{\rm STDP}(l)=S_{j}(l)\big[W_{{\rm pre}}+{\textstyle \int_{0}^{\infty}A_{+}W_{+}(y)S_{i}(l-y){\rm d}y\big]}+S_{i}(l)\big[W_{{\rm post}}+{\textstyle \int_{0}^{\infty}A_{-}W_{-}(y)S_{j}(l-y){\rm d}y\big],}
\end{equation}
where $ W_{\rm pre} $ and $ W_{\rm post} $ are respectively the constants of presynaptic and postsynaptic activity, $ A_{+} $ and $ A_{-} $ characterize the extent to which synaptic changes depend on the current synaptic weights. $ W_{+} $ and $ W_{-} $ are respectively the time windows of the long-term potentiation (LTP) and the long-term depression (LTD) processes, and they satisfy $ \int_{0}^{\infty}W(l){\rm d}l=1 $.
With these notations, each tuple $ \langle W_{\rm pre}, W_{\rm post}, A_{+}(w_{ij}), A_{-}(w_{ij}) \rangle $ defines an STDP learning rule.

\section{Method}
In this paper, we develop a spiking-based variational policy gradient (SVPG) method with R-STDP and an RWTA network. In this section, we first introduce the formulation of the policy and the network. Then we derive the parts of the method respectively for inference and optimization.

\subsection{Network Structure}
The key demand in network design is the capability for probabilistic inference so that the relationship between local optimization and global target can be built. It has been shown that WTA circuits with the STDP learning rule can perform probabilistic inference and learning \cite{Guo2019}. Here we consider rate coding -- each input (state) neuron represents a normalized state element; each output (action) state represents the probability of corresponding action being selected; each hidden WTA circuit encodes the probability distribution of a hidden variable.

\begin{wrapfigure}{R}{0.66\linewidth}
	\vskip -2mm
	\centering
	\includegraphics[width=0.86\linewidth]{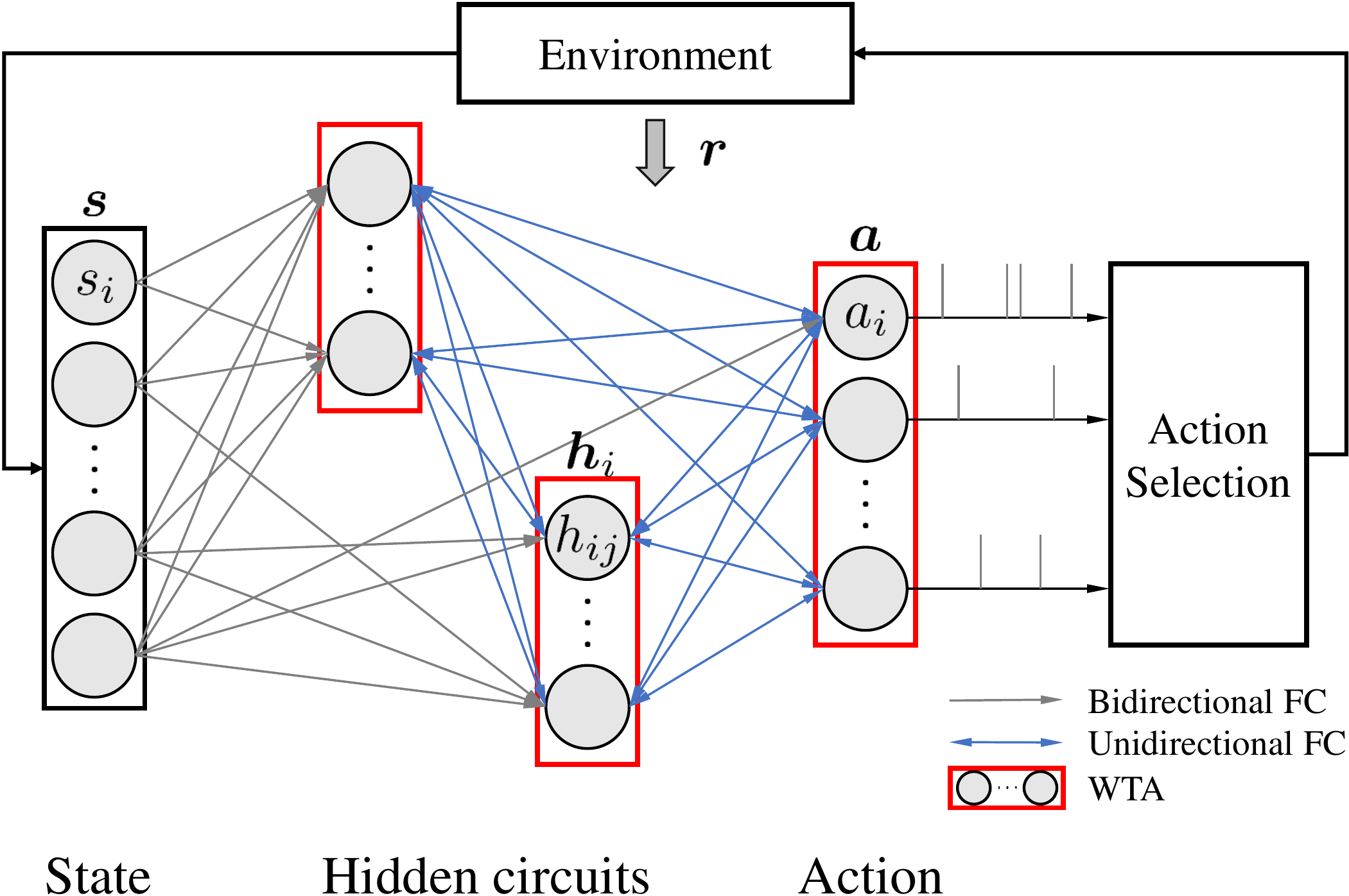}
	\caption{RWTA Network structure.}
	\label{fig:networkstructure}
	\vskip -3mm
\end{wrapfigure}



The RWTA network structure is shown in Figure \ref{fig:networkstructure}. Red rectangles represent WTA circuits. State neurons are denoted as $ s_i $ ($ i=1,\dots,d_s $); action neurons are denoted with $ a_i $ ($ i=1,\dots,d_a $); the $ j $-th neuron in the $ i $-th hidden circuit is denoted with $ h_{ij} $ ($ i=1,\dots,n_h $, $ j=1,\dots,d_h $). Here $ d_s $, $ d_h $, and $ d_a $ are the corresponding dimensions, and $ n_h $ is the number of hidden groups. ``FC'' means ``fully connected''.

Note that similar circuits have been investigated in \cite{Guo2019}. However, their structure is constrained to trees of circuits, while ours allows arbitrary connections and so covers any recurrent structures (regardless of layer number and layer size).


At each spike time step, each neuron has two properties: firing probability $ q\in [0, 1]$ and a binary firing status $ v\in \{0, 1\}$. For example of notations, the properties of neuron $ h_{ij} $ are $ q_{h_{ij}} $ and $ v_{h_{ij}} $. We use bold symbols to denote vectors. For example, the firing probability of WTA circuit $ h_i $ is
$ \boldsymbol{q}_{h_i}:=[q_{h_{i1}}, \dots,q_{h_{i d_h}}]^{\rm T} $,
where $ [\cdot]^{\rm T} $ is the transpose operation. 
For conciseness, we denote $ \boldsymbol{q}_{h}:= [\boldsymbol{q}_{h_1}^{\rm T}, \dots, \boldsymbol{q}_{h_{n_h}}^{\rm T}]^{\rm T}$, $ \boldsymbol{v}_{h}:= [\boldsymbol{v}_{h_1}^{\rm T}, \dots, \boldsymbol{v}_{h_{n_h}}^{\rm T}]^{\rm T}$, and then $ \boldsymbol{q}=[\boldsymbol{q}_{h}^{\rm T}, \boldsymbol{q}_{a}^{\rm T}, \boldsymbol{q}_{s}^{\rm T}]^{\rm T} $, $ \boldsymbol{v}=[\boldsymbol{v}_{h}^{\rm T}, \boldsymbol{v}_{a}^{\rm T}, \boldsymbol{v}_{s}^{\rm T}]^{\rm T} $.
The total number of neurons is $N=n_h d_h+d_a+d_s$. The synapse weights between all the neurons are altogether parameterized as $ \boldsymbol{W}\in \mathbb{R}^{N\times N} $. The self-activation parameters are $ \boldsymbol{b}\in\mathbb{R}^{N} $. The aforementioned parameters of a policy, $ \theta $, is thus the combination of $ \theta=\langle\boldsymbol{W}, \boldsymbol{b}\rangle $.
Note that different network structures can be represented by constraining certain parts of $ \boldsymbol{W} $ to zero.

\subsection{Policy Inference}
We consider the RL policy to be a probability distribution over the action space, and define it using an energy function $ E(\boldsymbol{v}) $:
\begin{gather}
	\pi(\boldsymbol{v}_{a}|s)={\textstyle \sum_{\boldsymbol{v}_{h}}p(\boldsymbol{v}_{a},\boldsymbol{v}_{h}|s)}, \label{eq_policy_marginal}
	\\
	p(\boldsymbol{v}_{a},\boldsymbol{v}_{h}|s):={\textstyle \frac{1}{Z(s)}}\exp\{E(\boldsymbol{v})\},\qquad E(\boldsymbol{v}):=\boldsymbol{v}^{{\rm T}}\boldsymbol{W}\boldsymbol{v}+\boldsymbol{b}^{{\rm T}}\boldsymbol{v},
	\label{eq_policy_energy}
\end{gather}
where $ Z(s) $ is the normalization $ Z(s)=\sum_{\boldsymbol{v}_{h}', \boldsymbol{v}_{a}'}\exp\{E(\boldsymbol{v}')\} $. Given a state $ s $, first the encoding $ \boldsymbol{v}_s $ is generated (see experiment section for details), and then the action distribution is calculated as the marginal distribution of $ \boldsymbol{v}_{a} $. Although the energy function is linear, the normalization operation enables representation of complex distributions \cite{heess_actor-critic_2013}.

This policy representation is computable in principle. However, when there is a large number of hidden neurons, the calculation can be intractable. To address this problem, we adopt mean-field inference to derive a tractable method. We will then show that the mean-field inference can be realized with the RWTA network.

\subsubsection*{Policy Mean Field Inference}
First, we use a variational distribution $ \hat{p}(\boldsymbol{v}_{a}, \boldsymbol{v}_{h}|s) $ to approximate $ p(\boldsymbol{v}_{a}, \boldsymbol{v}_{h}|s) $, and assume that the states of all circuits are independent from each other. This allows decomposition of $ \hat{p} $:
$ \hat{p}(\boldsymbol{v}_{a}, \boldsymbol{v}_{h}|s):=\hat{p}(\boldsymbol{v}_{a}|s) \hat{p}(\boldsymbol{v}_{h_1}|s) \cdots \hat{p}(\boldsymbol{v}_{h_{n_h}}|s) $,
where $ \hat{p}(\boldsymbol{v}_{h_1}|s) := \boldsymbol{q}_{h_1}^{\rm T}\boldsymbol{v}_{h_1} $, $ \dots $, $ \hat{p}(\boldsymbol{v}_{a}|s) = \boldsymbol{q}_{a}^{\rm T}\boldsymbol{v}_{a} $.

Second, we use the KL divergence to measure the difference between $ \hat{p} $ and $ p $: $ D_{{\rm KL}}(s)\overset{\cdot}{=}D_{{\rm KL}}[\hat{p}(\boldsymbol{v}_{a},\boldsymbol{v}_{h}|s)\Vert p(\boldsymbol{v}_{a},\boldsymbol{v}_{h}|s)] $.
By letting $ \partial D_{{\rm KL}}(s)/\partial q_i=0 $ for each $ q_i $ (the $ i $-th element of vector $ \boldsymbol{q} $), we get the mean-field inference function \cite{DBLP:journals/ftml/WainwrightJ08}:
\begin{equation}\label{ep_KL_derivation}
	\begin{gathered}q_{i}={\textstyle \frac{1}{Z(q_{G(i)})}\exp\{\boldsymbol{w}_{{\rm row},i}^{{\rm T}}\boldsymbol{q}+\boldsymbol{w}_{{\rm col},i}^{{\rm T}}\boldsymbol{q}+b_{i}\},}
	\end{gathered}
\end{equation}
where $ Z(q_{G(i)})={\textstyle \sum_{j\in G(i)}}\exp\{\boldsymbol{w}_{{\rm row},j}^{{\rm T}}\boldsymbol{q}+\boldsymbol{w}_{{\rm col},j}^{{\rm T}}\boldsymbol{q}+b_{j}\} $, $ i = 1,\dots,(n_h d_h +d_a) $, $ G(i) $ is the set of indices of the neurons in the same circuit as neuron $ i $, and $ \boldsymbol{w}_{{\rm row},i} $ and $ \boldsymbol{w}_{{\rm col},i} $ are respectively the $ i $-th row and column of matrix $ \boldsymbol{W} $ (in the shape of a column vector), which corresponds to the synapses connected to neuron $i$. $ b_i $ is the $ i $-th element in vector $ \boldsymbol{b} $.

So far we derived an approximate policy representation -- by finding a $ \boldsymbol{q} $ that satisfies Eq.(\ref{ep_KL_derivation}), we can get the $ \boldsymbol{q}_a $ within it, which is the policy distribution we need.
Notice that Eq.(\ref{ep_KL_derivation}) can been seen as an iteration process by regarding the $ \boldsymbol{q} $ on the right side as constant. So an iterative method can be used to find $ \boldsymbol{q} $ -- initialize $ \boldsymbol{q} $ with random numbers, then repeat updating it with Eq.(\ref{ep_KL_derivation}) until numeric convergence. Although there has not been a theoretical guarantee of convergence, the iteration converges in our experiments.

\subsubsection*{Policy Inference Based on RWTA Network}
Now we show that this iterative method for inference can be realized with the RWTA network. Since the inhibitory neuron in a WTA circuit produces an overall firing rate $\hat{\rho}$ \cite{Guo2019}, a natural link is to let firing probabilities encode $ \rho_{i}(l)=\hat{\rho}q_{i} $. This transforms Eq.(\ref{ep_KL_derivation}) into (note that $ 1/Z(q_{G(i)})=\exp\{-\log[Z(q_{G(i)})]\} $):
\begin{equation}\label{eq_trans_1}
	\rho_{i}=\hat{\rho}\exp\{\boldsymbol{w}_{{\rm row},i}^{{\rm T}}\boldsymbol{q}+\boldsymbol{w}_{{\rm col},i}^{{\rm T}}\boldsymbol{q}+b_{i}-\log{\textstyle \sum_{j\in G(i)}\exp\{\boldsymbol{w}_{{\rm row},j}^{{\rm T}}\boldsymbol{q}+\boldsymbol{w}_{{\rm col},j}^{{\rm T}}\boldsymbol{q}+b_{j}\}\}} .
\end{equation}
By replacing $ w_j $ with the synaptic weights $ w_{ij} $ in Eq.(\ref{eq_potential}), and designing $ \kappa(y) $ such that $ \int_{0}^{\infty}\kappa(y){\rm d}y =1/\hat{\rho} $, we can use the potential $ u(l) $ to replace the firing probability $ q $. This results in the following spike-based inference function:
\begin{equation}\label{eq_trans_2}
	\begin{gathered}\rho_{i}(l)=\hat{\rho}\exp\{u_{i}(l)-\log{\textstyle \sum_{j\in G(i)}\exp(u_{j}(l))\}},\\
		u_{i}(l)={\textstyle \sum_{j\in N(i)}w_{ij}{\textstyle \int_{0}^{\infty}\kappa(y)S_{ij}(l-y){\rm d}y+b_i}}.
	\end{gathered}
\end{equation}
This function is biologically plausible since the firing probability of each neuron is determined by its potential, which is determined by the spike trains from neighboring neurons. The only difference from the LIF model Eq.(\ref{eq_firing}, \ref{eq_potential}) is that the potential threshold $ u_{\rm th} $ is replaced by a logarithm item, which can be made by introducing lateral inhibition neurons \cite{kappel2018dynamic}.

\subsection{Policy Optimization}
As has been introduced in section \ref{sec_RL_PG}, the key to policy gradient is to derive the differential of $ \log \pi_\theta(a_t|s_t) $. This means to derive the differential of each $q_i$ regarding each parameter $ w_{jk} $ and $b_j$, using Eq.(\ref{ep_KL_derivation}). In this section, we first derive the precise differential, then show how R-STDP can represent its first-order approximation.

\subsubsection*{Precise Differential}

\begin{theorem}\label{Th1}
	The precise differential of $\boldsymbol{q}_{ha}$ with reference to the synaptic weight $ w_{jk}$ and the self-activation parameter $b_{j}$ is as follows:
	\begin{equation}\label{eq_diff_matrix}
		\begin{gathered}{\textstyle \frac{\partial\boldsymbol{q}_{ha}}{\partial w_{jk}}}=\boldsymbol{M}\big(\boldsymbol{U}_{jk}+\boldsymbol{U}_{kj}\big)\boldsymbol{q}+\boldsymbol{M}(\boldsymbol{W}+\boldsymbol{W}^{{\rm T}}){\textstyle \frac{\partial\boldsymbol{q}}{\partial w_{jk}}},
		\quad
		{\textstyle \frac{\partial\boldsymbol{q}_{ha}}{\partial b_{j}}}=\boldsymbol{M}\boldsymbol{b}+\boldsymbol{M}(\boldsymbol{W}+\boldsymbol{W}^{{\rm T}}){\textstyle \frac{\partial\boldsymbol{q}}{\partial b_{j}}},\\
			\boldsymbol{M}={\rm diag}(\boldsymbol{q}_{ha})[-\boldsymbol{G}_{ha}{\rm diag}(\boldsymbol{q})+\boldsymbol{D}_{{\rm sel}}],
		\end{gathered}
	\end{equation}
	where $ \boldsymbol{G}_{ha} $ is a $ (n_{h}d_{h}+d_{a})\times N $ logical matrix of which $ 1 $ elements indicate the two neurons (column index and row index) are in the same circuit, $ \boldsymbol{D}_{{\rm sel}} $ is a logical matrix that selects the first $ (n_h d_h +d_a) $ elements in a vector with length $ N $, i.e., $ \boldsymbol{D}_{{\rm sel}}=\begin{bmatrix}\boldsymbol{I}_{(n_{h}d_{h}+d_{a})} & \boldsymbol{O}_{(n_{h}d_{h}+d_{a})\times d_{s}}\end{bmatrix} $,
	and $ \boldsymbol{U}_{jk} $ is a $ N \times N $ logical matrix with only the $ jk $-th element being $ 1 $.
\end{theorem}
The proof is given in Appendix \ref{PT1}.
Theorem 1 reveals that
$ {\partial\boldsymbol{q}_{ha}}/{\partial w_{jk}} $
can be calculated by solving the matrix equations in Eq.(\ref{eq_diff_matrix}). Specifically, the calculation includes the matrix pseudo-inverse of $ \boldsymbol{M}(\boldsymbol{W}+\boldsymbol{W}^{{\rm T}}) $, whose shape is $ (n_h d_h + d_a)\times N $. So the calculation complexity is more than $O((n_h d_h + d_a)^3)$, which can be intractable for large neuron numbers.
More importantly, this solution does not reveal the link between local parameter updates and the global RL target, and is not biologically plausible.

\subsubsection*{First-Order Approximation of the Differential}
Below we derive a first-order approximation to the differential, and show how it can be implemented by an R-STDP learning rule. As will be shown in the experiment section, this approximation does not hinder the network from solving complex RL tasks.

The main idea for the approximation is to treat the $ \boldsymbol{q} $ on the right side of the inference function Eq.(\ref{ep_KL_derivation}) as a constant with reference to $ w_{jk} $ and $ b_{j} $. By doing so, the differential only considers the last iteration in the inference process, where the status of each neuron is only affected by its neighboring neurons. Thus the differential only requires the properties of neighboring neurons, which enables the link to local learning rules like R-STDP.

To get the policy gradient $ \nabla \log \pi(a_t|s_t) $, $ {\partial\log(q_{i})} / {\partial w_{jk}} $ and $ {\partial\log(q_{i})} / {\partial b_{j}} $ are required. We present the results in the following Theorem \ref{Th2}. The proof is given in Appendix \ref{PT2}.
\begin{theorem}\label{Th2}
	The approximate differentials of firing rate $q_{i}$ with respect to $w_{jk}$ and $b_j$ in the matrix form are:
	\begin{equation}\label{eq_diff_approx_w_matrix}
		\begin{aligned}\textstyle\frac{\partial\log(q_{i})}{\partial\boldsymbol{W}}= & (\boldsymbol{U}_{i:}{\rm diag}(\boldsymbol{q})+{\rm diag}(\boldsymbol{q})\boldsymbol{U}_{:i})-{\rm diag}(\boldsymbol{q})(\boldsymbol{U}_{G(i):}+\boldsymbol{U}_{:G(i)}){\rm diag}(\boldsymbol{q}),\\
			\textstyle\frac{\partial\log(q_{i})}{\partial\boldsymbol{b}}= & \boldsymbol{u}_{i}-{\rm diag}(\boldsymbol{q})\boldsymbol{u}_{G(i)},
		\end{aligned}
	\end{equation}
	where $ i\in\{1,\dots,(n_h d_h + d_a)\} $, $ \boldsymbol{U} $ is a $ N\times N $ logical matrix and $ \boldsymbol{u} $ is a length-$ N $ logical vector, whose subscripts indicates the positions of elements with value $ 1 $. $ G(i) $ is the set of indices of neurons in the same circuit as neuron $ i $; ``$ : $'' means the entire row/column.
\end{theorem}

With this result, when the firing state of the network is $ \boldsymbol{v} $, the corresponding REINFORCE policy gradient is
\begin{equation}\label{eq_gradient}
	\nabla J(\pi)={\textstyle \sum_{t}}\gamma^{t}r_{t}[{\textstyle \sum_{i=1}^{n_{h}}}\boldsymbol{v}_{h_{i}}^{{\rm T}}\nabla(\log\boldsymbol{q}_{h_{i}})+\boldsymbol{v}_{a}^{{\rm T}}\nabla(\log\boldsymbol{q}_{a})] ,
\end{equation}
where $ \nabla \log\boldsymbol{q}_{h_{i}} $ and $ \nabla \log\boldsymbol{q}_{a} $ are respectively the vectors of $ \nabla \log q_{h_{ij}} $ and $ \nabla \log q_{a_i} $. So far we can learn the RWTA network by the REINFORCE algorithm to optimize the RL target.

\subsubsection*{Policy Optimization Based on R-STDP}
Notice that in the policy gradient Eq.(\ref{eq_diff_approx_w_matrix}), the terms of $ w $ only involve pre- and post-synaptic neurons, and the terms of $ b $ only involve connected neurons, which is similar to the R-STDP rules. Now we show how it can be linked to R-STDP.

Recall an R-STDP rule is determined by $ \langle W_{\rm pre}, W_{\rm post}, A_{+}(w_{ij}), A_{-}(w_{ij}) \rangle $. By assuming a large enough spike step number in each RL step, we have that the averaged number of spikes in the spiking train $S_i$ approximates the firing probability $\rho_i$. That is $\mathbb{E}[S_i(l)]=\rho_i$ and $\mathbb{E}[\int_{0}^{\infty}A_{+}(w_{ij})S_{i}(l-y){\rm d}y]=A_{+}(w_{ij})\rho_{i}$. By performing the following settings of STDP (element-wise notation for the synapse between neuron $i$ and $j$):
\begin{equation}\label{eq_STDP_transformation}
	\langle W_{{\rm pre}},W_{{\rm post}},A_{+}(w_{ij}),A_{-}(w_{ij})\rangle=\langle v_{i},v_{j},-1/\hat{\rho},-1/\hat{\rho}\rangle,
\end{equation}
we get the following R-STDP rule:
\begin{equation}\label{eq_RSTDP}
	\mathbb{E}[R(l){\rm STDP}(l)]=R[\rho_{j}(v_{i}-\rho_{i}/\hat{\rho})+\rho_{i}(v_{j}-\rho_{j}/\hat{\rho})].
\end{equation}
Then, for the policy gradient we have just derived (Eq.\ref{eq_diff_approx_w_matrix}, \ref{eq_gradient}), there is
\begin{equation}
	{\textstyle \frac{\partial J(\pi)}{\partial w_{ij}}=}{\textstyle \sum_{t}}\gamma^{t}r_{t}[q_{i}(v_{j}-q_{j})+q_{j}(v_{i}-q_{i})],
\end{equation}
which is the same as Eq.(\ref{eq_RSTDP}) (despite a constant $\hat{\rho}$). This shows that an R-STDP characterized by Eq.(\ref{eq_STDP_transformation}) can represent the policy gradient on synapse weights.
As for the self-activation parameter $b$, a similar transformation exists as $ {\rm \Delta}b(l)=v_{i}+{\textstyle \int_{0}^{\infty}-(1/\hat{\rho})S_{i}(l-y){\rm d}y} $.

So far we get a variational policy gradient method where inference and optimization are implemented with the spiking RWTA network and an R-STDP rule. We name it \textit{SVPG}.

Note that although our method is derived from the REINFORCE algorithm, it can be adapted to many other policy gradient algorithms. Some implementation details, including adaptation to Actor-Critic algorithms, are provided in Appendix \ref{APPENDIX_SVPG_implementation_details}.

\section{Experiments}

\subsection{Task Design}
We selected two commonly used RL tasks: reward-based MNIST image classification (\textit{MNIST}) and Gym InvertedPendulum (\textit{GymIP}). MNIST is a one-step task featured by a high dimensional input ($28\times 28$). GymIP is a widely-used task in RL researches \cite{chung_reinforcement_2020,lovatto_analyzing_2019,Ozalp2020} featured by long-term decision sequences. Details of the task goals, reward settings, state processing, etc. are in Appendix \ref{APPENDIX_task_details}.

\begin{figure*}[t]
	\begin{minipage}[b]{0.46\linewidth}
		\begin{figure}[H]
			\centering
			\vspace{-3mm}
			\subfigure[Pole length]{
				\centering
				\includegraphics[width=0.45\linewidth]{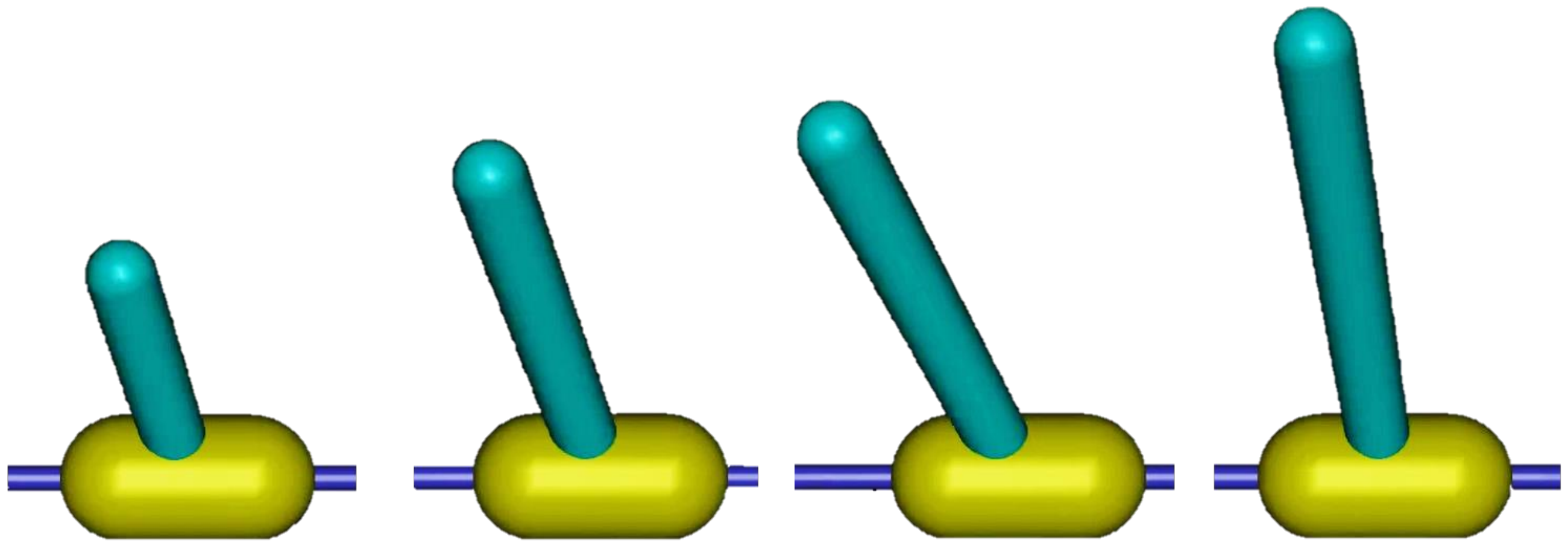}
				\label{fig_ip_length}
			}
			\subfigure[Pole thickness]{
				\centering
				\includegraphics[width=0.45\linewidth]{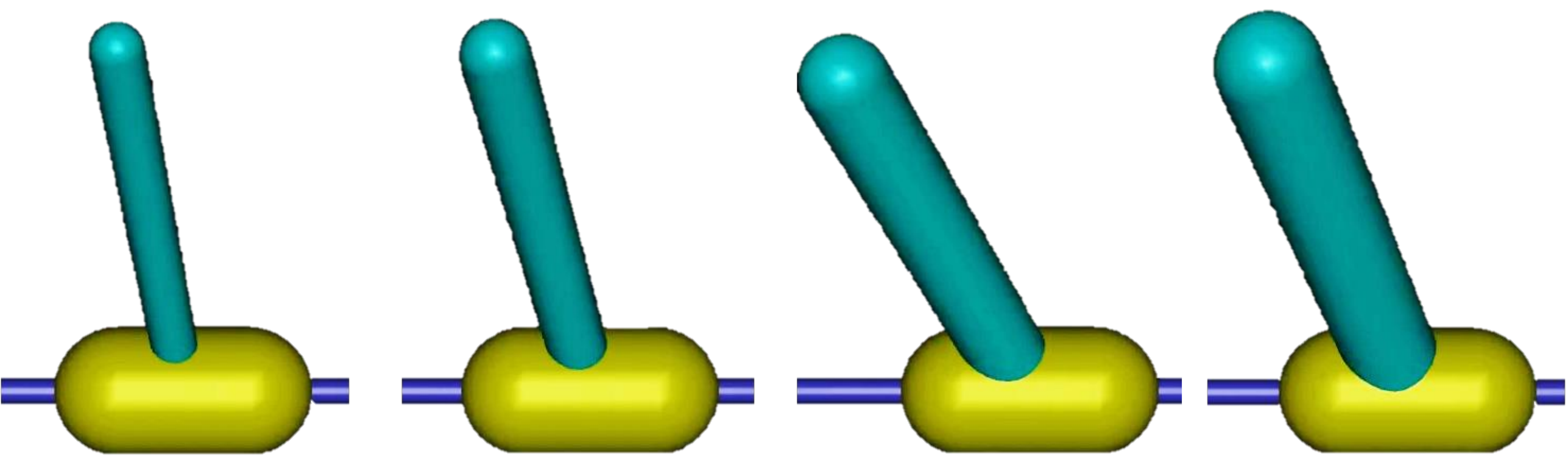}
				\label{fig_ip_thick}
			}
			\caption{Two types of changes to GymIP.}
			\label{fig:exp_ip_variations}
		\end{figure}
	\end{minipage}
	\hfill
	\begin{minipage}[b]{0.52\linewidth}
		\begin{table}[H]
			\begin{center}
				\resizebox{\linewidth}{6mm}{
					\begin{tabular}{cccccc}
						\toprule 
						& SVPG & BP & BPTT & EP & ANN2SNN\tabularnewline
						\midrule
						MNIST & 0.926$\pm$   0.001 & 0.933$\pm$   0.062 & 0.898$\pm$   0.067 & 0.971$\pm$   0.001 & 0.933$\pm$   0.062 \tabularnewline
						\midrule 
						GymIP & 199.87 $\pm$ 0.27 & 199.95 $\pm$ 0.13 & 199.96 $\pm$ 0.12 & N/A & 190.79 $\pm$ 27.64\tabularnewline
						\bottomrule
				\end{tabular} }
			\end{center}
			\caption{Zero-noise performances (mean$ \pm $std).}
			\label{tab:zero_noise_performances}
		\end{table}
	\end{minipage}
\end{figure*}

It has been shown that, compared to ANNs, SNNs trained with ANN2SNN methods can have better robustness to input and synapse weight noises \cite{patel2019improved,li_robustness_2020}, and that SNNs trained with BPTT can be more resilient to adversarial attacks \cite{sharmin_comprehensive_2019}. Therefore it is worth checking whether SVPG brings better robustness than ANNs and other SNN learning methods.
We test three types of disturbance -- input noise, network parameter noise, and GymIP environmental variation.
\textbf{1) Input noise} is independently added to each dimension of state observations. For MNIST, Gaussian, salt, salt\&pepper, and Gaussian\&salt noises are considered. For GymIP, Gaussian and uniform noises are considered. Some illustrations are in Appendix \ref{sec:MNIST_noise}.
\textbf{2) Network parameter noise} is independently added to each of the learnable parameters in the policy networks. Gaussian and uniform noises are considered. 
\textbf{3) Environment variations in GymIP}. For GymIP, the pendulum's length and thickness are modified during testing. In training the setting is <length=1.5, thickness=0.05>. In testing we change the length in the range of $ [0.16, 4.9] $ and thickness of $ [0.02, 0.30] $. Figure \ref{fig:exp_ip_variations} illustrates these variations.

\subsection{Compared Methods}
We select four representative learning methods. 1) \textit{ANN2SNN} with the methods from \cite{rueckauer_conversion_2017}. 2) \textit{BP} \cite{lecun_deep_2015} with the ReLU function for hidden layers, which is a conventional baseline ANN model using multilayer perceptrons architecture learning with error back-propagation. This serves as a usual approach in RL. 3) Fast sigmoid \textit{BPTT} (backpropagation through time) from \cite{zenke_superspike_2018}.  4) \textit{EP} (Equilibrium Propagation ) \cite{scellier_equilibrium_2017}, which is a biologically plausible algorithm with local learning rules, though not designed for SNNs. Note the EP is designed for supervised learning, so we only test it in MNIST using ground truths.
For all these methods, the number of hidden layers is set to  $ 1 $; the optimizer is set to stochastic gradient descent with zero momentum. For RL algorithms, we select REINFORCE \cite{sutton_reinforcement_2018} for the MNIST task and Advantage Actor-Critic \cite{mnih_asynchronous_2016} for the GymIP task. Hyper-parameters including the number of hidden neurons and RL discount factor $ \gamma $ are kept the same across methods. Note that the RWTA network in our method is fully-connected, which means it has more learnable parameters when having the same number of hidden neurons. To make the comparison fair we introduce a \textit{SVPG-shrink} with fewer hidden WTA circuits, so that the number of learnable parameters are similar in all the models. 
Details are provided in Appendix \ref{sec:Appendix_alg_details}.

\begin{figure*}
	\begin{minipage}[b]{0.78\linewidth}
		\vspace{-5mm}
		\begin{minipage}[b]{0.49\linewidth}
			\begin{figure}[H]
				\centering
				\subfigure[\scriptsize Input salt noise]{\raggedright\includegraphics[height=18mm]{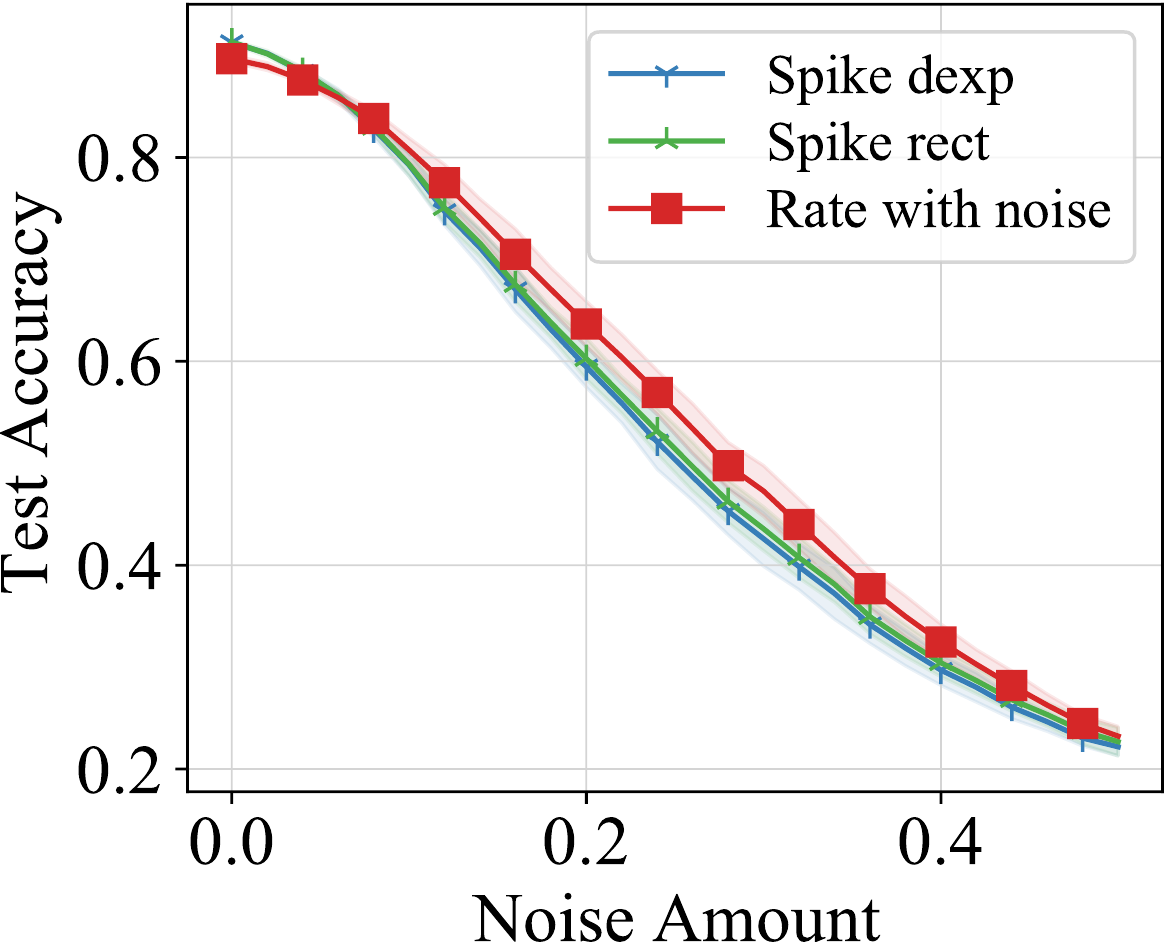}}
				\hfill
				\subfigure[\scriptsize Net Gaussian noise]{\raggedright\includegraphics[height=18mm]{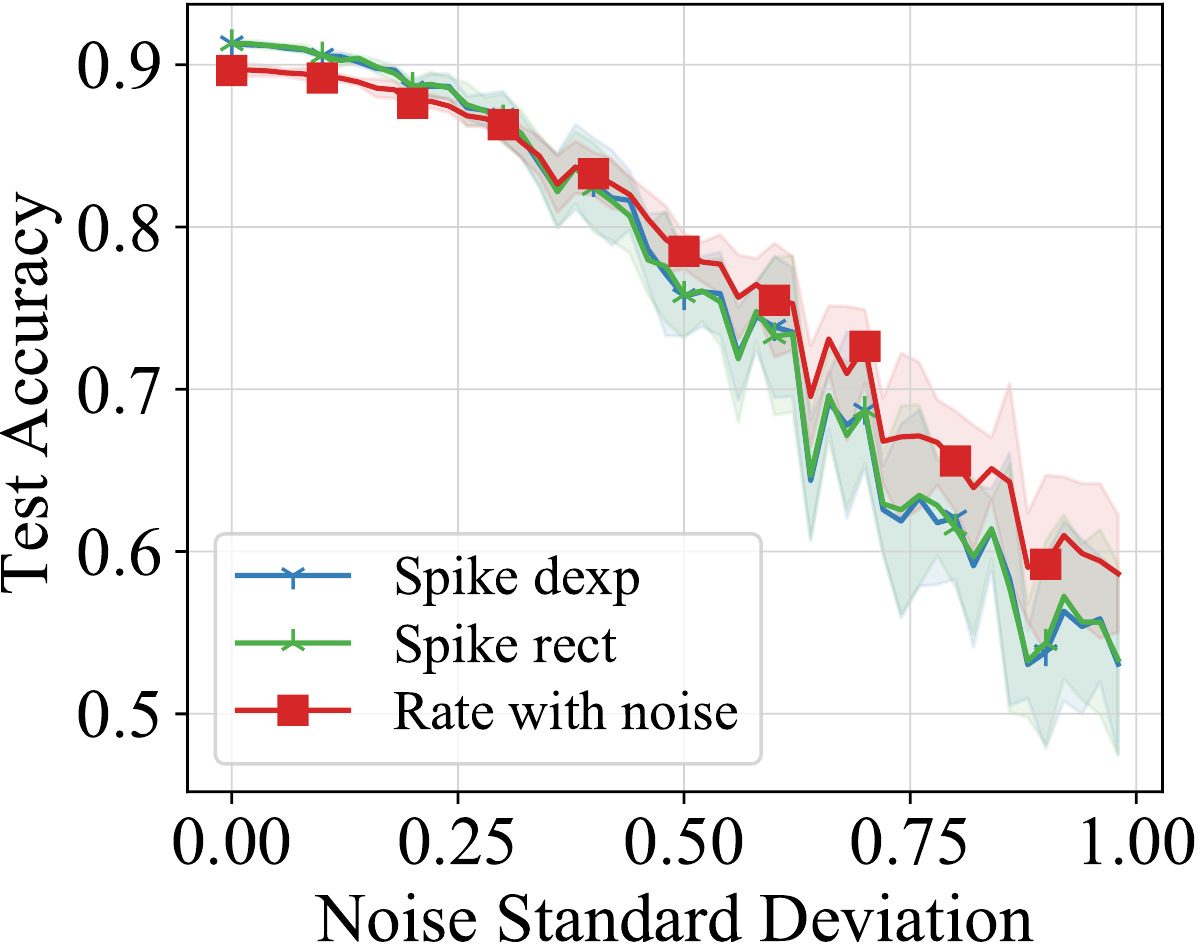}}
				\caption{SVPG: spike/rate coding.\label{fig:SVPG_compare}}
			\end{figure}
		\end{minipage}
		\hfill
		\begin{minipage}[b]{0.50\linewidth}
			\begin{figure}[H]
				\centering
				\subfigure[\scriptsize Input Gaussian noise\label{fig:mnist_input}]{\raggedright\includegraphics[height=19.5mm]{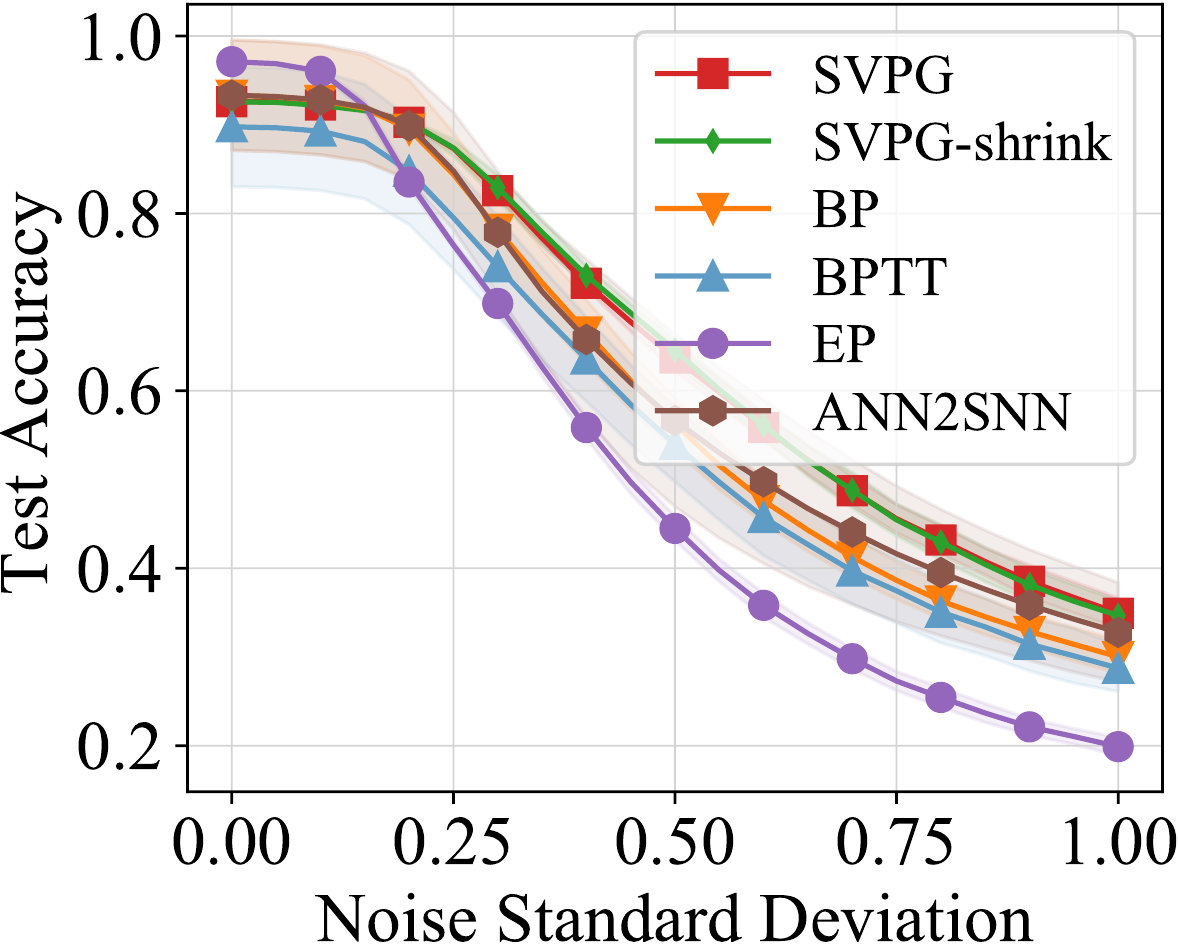}}
				\hfill
				\subfigure[\scriptsize Net Gaussian noise\label{fig:mnist_net}]{\raggedright\includegraphics[height=19.3mm]{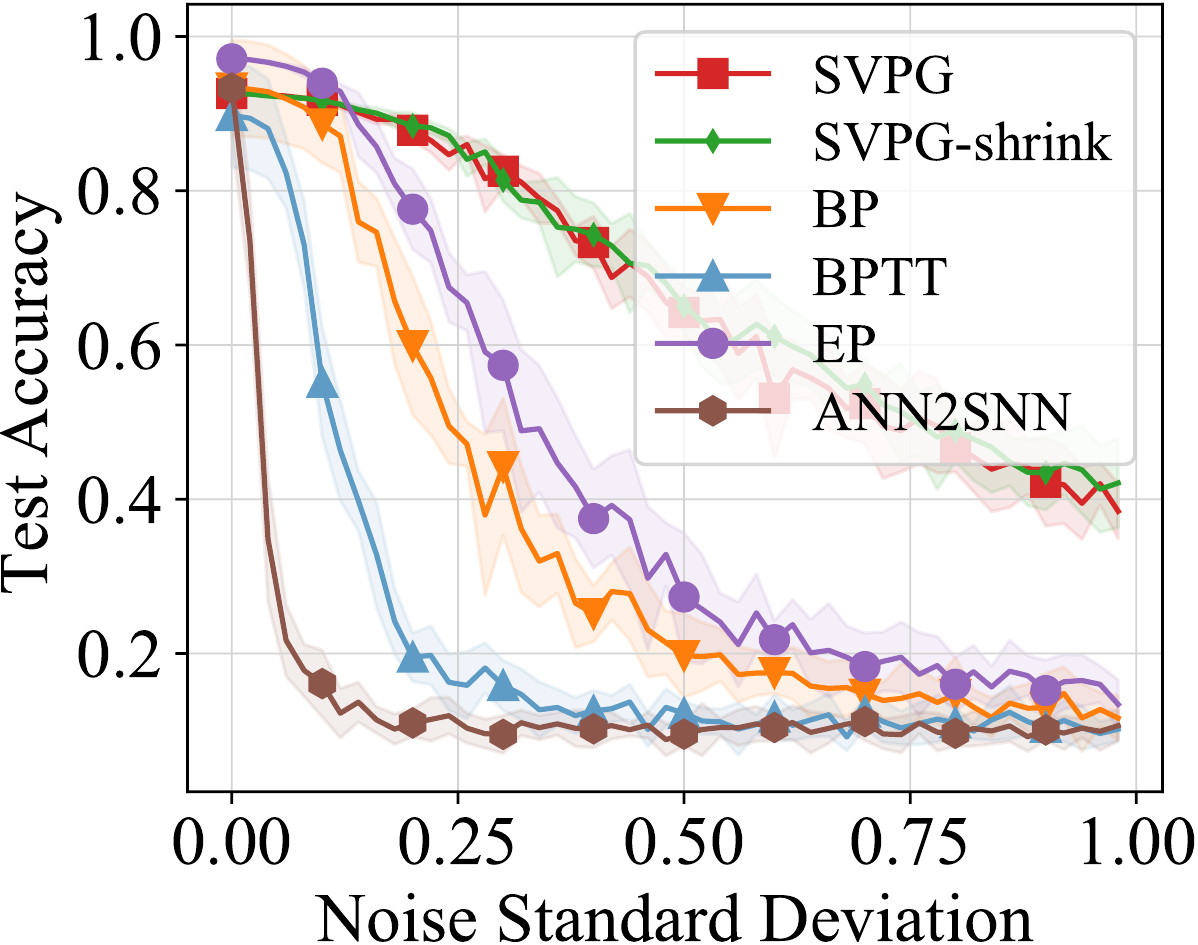}}
				\caption{MNIST: Robustness.}
				\label{fig:robust_MNIST}
			\end{figure}
		\end{minipage}
		
		\begin{minipage}[b]{0.99\linewidth}
			\begin{figure}[H]
				\subfigure[\scriptsize Input uniform noise\label{fig:gymip_input}]{\raggedright\includegraphics[height=19.5mm]{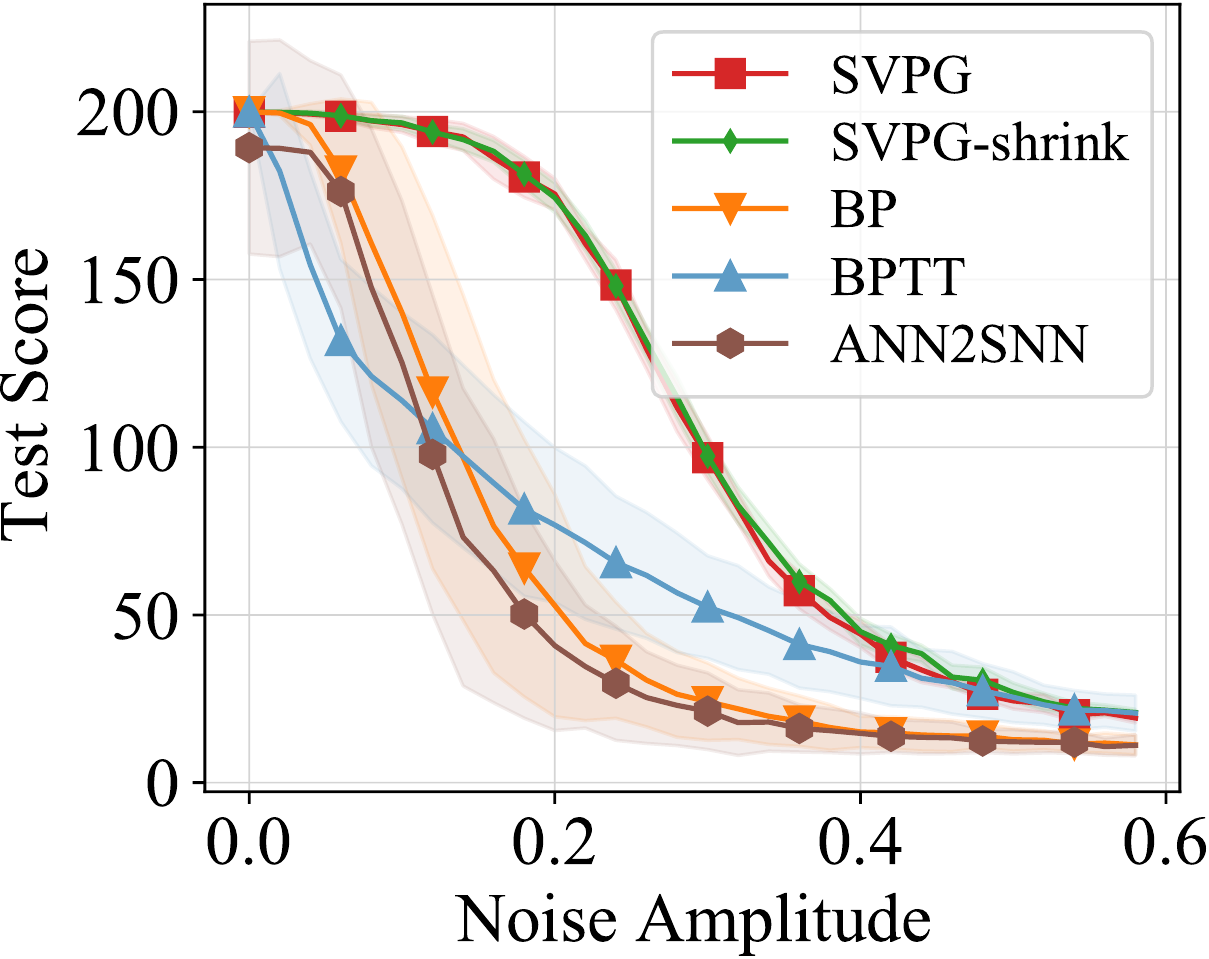}}
				\hfill
				\subfigure[\scriptsize Net uniform noise\label{fig:gymip_net}]{\raggedright\includegraphics[height=19.5mm]{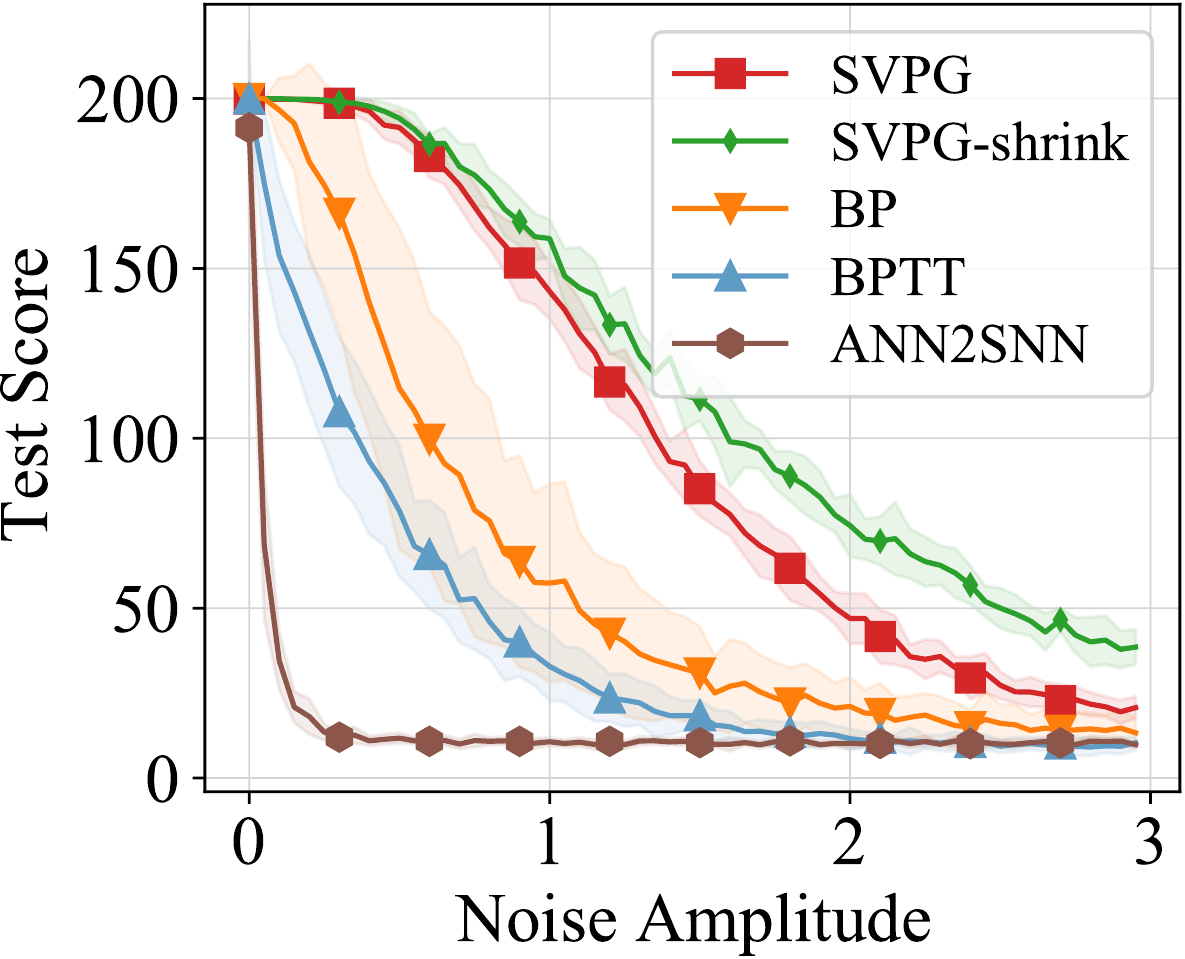}}
				\hfill
				\subfigure[\scriptsize Pendulum length\label{fig:gymip_length}]{\raggedright\includegraphics[height=19.5mm]{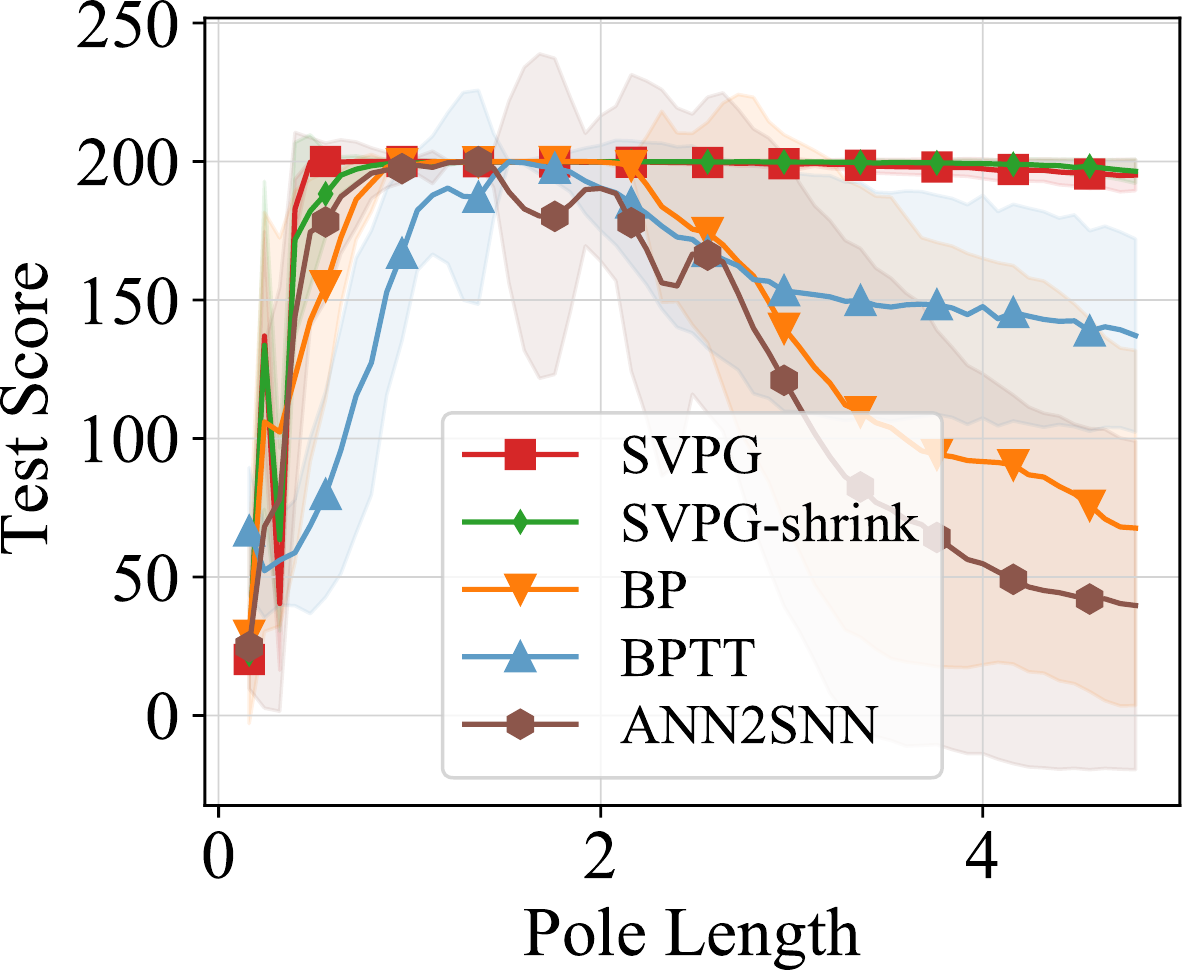}}
				\hfill
				\subfigure[\scriptsize Pendulum thickness\label{fig:gymip_thick}]{\raggedright\includegraphics[height=19.5mm]{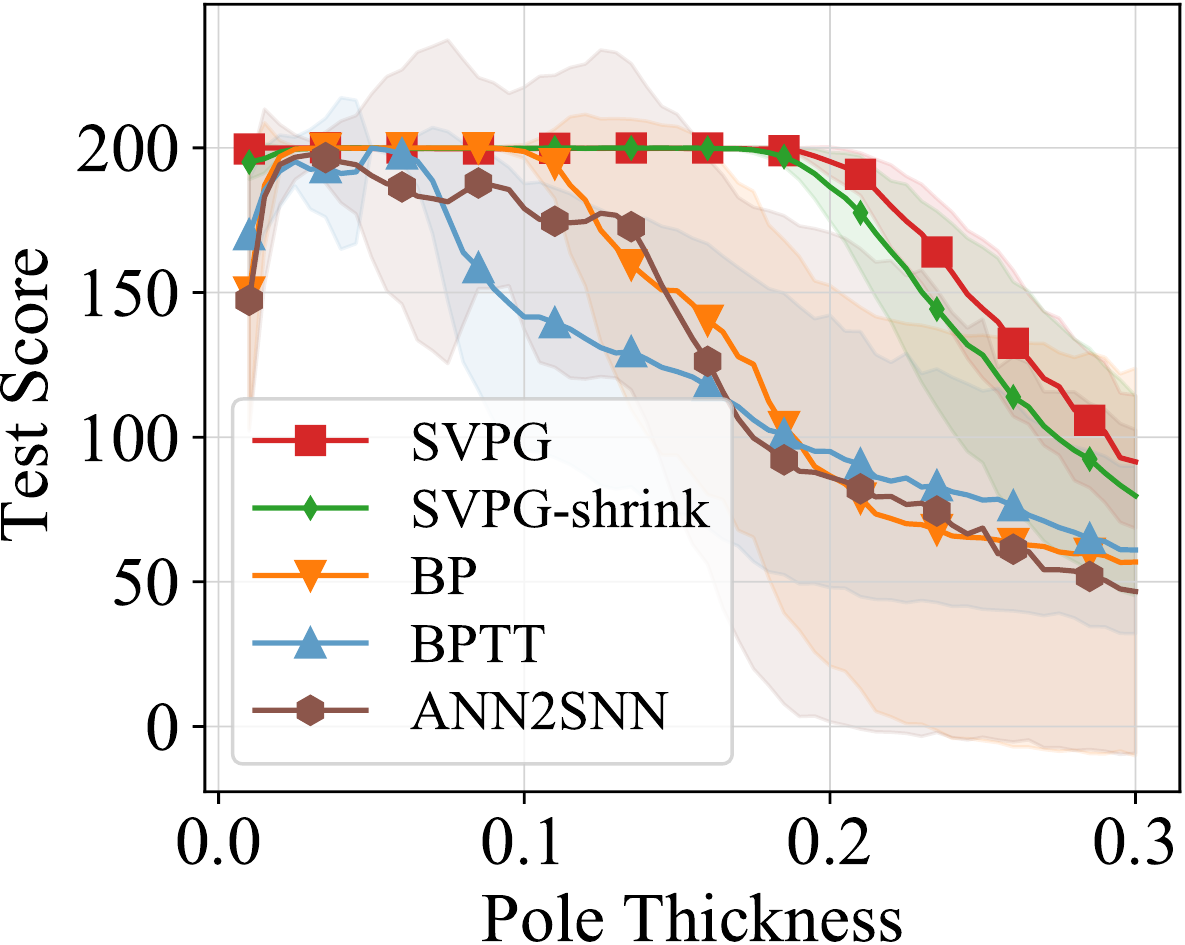}}
				\vspace{-4mm}
				\caption{GymIP: Robustness.\label{fig:robust_GymIP}}
			\end{figure}
		\end{minipage}
	\end{minipage}
	\unskip\,\textcolor[RGB]{192, 192, 192}{\vrule width 0.4pt} \,
	\begin{minipage}[b]{0.20\linewidth}
		\raggedright
		\begin{figure}[H]
			\subfigure[\scriptsize Net weights\label{fig:M_net_weight_dist}]{\raggedright\includegraphics[width=0.9\linewidth]{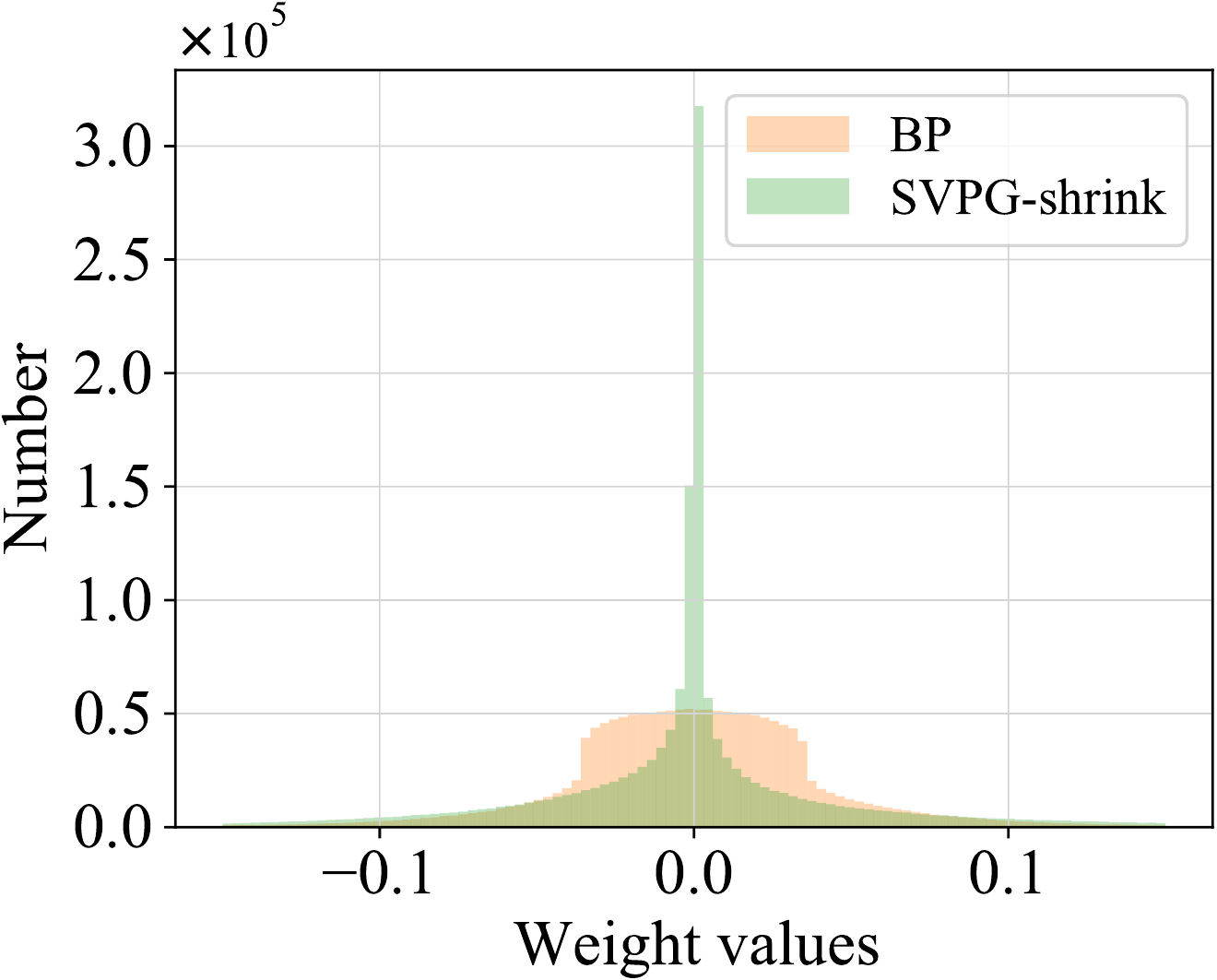}}
			
			\subfigure[\scriptsize Noise responses\label{fig:noise_response_distribution}]{\raggedright\includegraphics[width=0.96\linewidth]{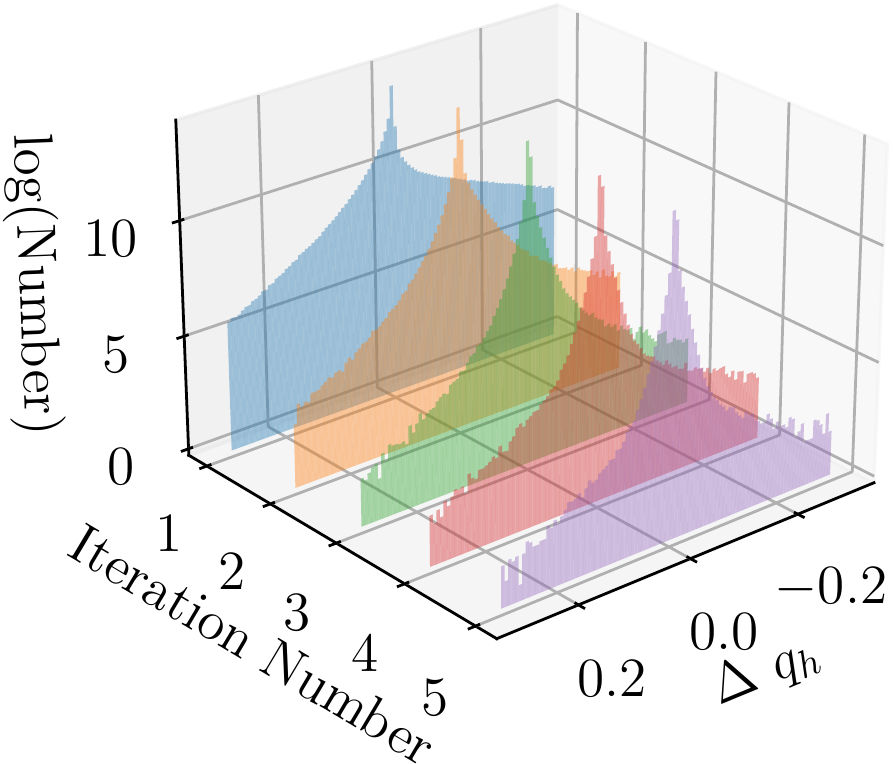}}
			\caption{\\Visualizations.}
		\end{figure}
	\end{minipage}
	\vskip -3mm
\end{figure*}

\subsubsection*{SVPG: spike coding versus rate coding}
The spike train simulation in SVPG is computationally expensive for common GPUs. With Eq.(\ref{ep_KL_derivation}) and Eq.(\ref{eq_gradient}), SVPG can also be implemented with rate-coded neuron states, which is more computationally efficient. To reproduce noises caused by spike-train simulation, we add Gaussian noise to rate coding after each iteration of Eq.(\ref{ep_KL_derivation}). This results in the ``\textit{Rate with noise}'' implementation.

To check whether rate coding alters the performance, we compare it with spike coding in the MNIST task. We consider two different STDP windows -- double exponential and rectangle -- and name them as ``\textit{Spike dexp}'' and ``\textit{Spike rect}''. As in Figure \ref{fig:SVPG_compare} and Appendix \ref{sec:Additional_comparison_implementation}, the three implementations generate similar curves, which indicates the rate coding implementation can be used as a replacement for spike coding. In the following experiments, we adopt the rate coding implementation.

\subsection{Results}
\subsubsection{Training Performances}
We perform 10 independent training trials (each with a different seed for the random number generator; each with the same number of episodes) for each method. We regularly save checkpoints during training and select the ones with the best zero-noise testing performance as the training results, which are summarized in Table \ref{tab:zero_noise_performances}. 1) In MNIST, SVPG achieves an accuracy close to BP and outperforms other RL-based methods. Note that although EP achieves high accuracy, it is trained using supervised signals, i.e., does not suffer from the distribution shift in RL. 2) In GymIP, SVPG achieves a near-optimal performance which is close to BP and BPTT. These results indicate that SVPG is able to solve both image-input and long-term decision tasks.

\subsubsection{Robustness}
We test the learned policies with different kinds of noises. To evaluate the performance, in MNIST, the testing dataset is used; in GymIP, 100 independently sampled episodes are used. We average the results across 10 training trials with different seeds, and present the standard deviations as shaded areas in the figures.

	1) \textit{Input noise}. The results of Gaussian noise in MNIST and uniform noise in GymIP are presented in Figure \ref{fig:mnist_input}, \ref{fig:gymip_input}. Others are in Appendix Figure \ref{fig:M_I_noise}, \ref{fig:G_I_noise}. For all the tested input noises, the performance of SVPG degrades the slowest as the amount/strength of noises increases. For 0.2 uniform noise in the GymIP task, SVPG even achieves more than twice the accuracy of all the compared methods.

	2) \textit{Network parameter noise}. The results of Gaussian noise in MNIST and uniform noise in GymIP are in Figure \ref{fig:mnist_net}, \ref{fig:gymip_net}. Other results are in Appendix Figure \ref{fig:M_W_noise}, \ref{fig:G_W_noise}. SVPG achieves slower degradation of performance as the amplitude of noises increases and has a better performance than all compared methods when the strength of noises is large.

	3) \textit{Environment variations in GymIP}.
	Results on variations in the pendulum's length and thickness are in Figure \ref{fig:gymip_length}, \ref{fig:gymip_thick}. Other variation results are in Appendix Figure \ref{fig:robust_env_GIP}. When the shape of the pendulum deviates from the one in training, the performance of all the compared methods degrades. However, the performance of SVPG degrades much slower than the baselines. Especially, when changing the pendulum's length from 1.5 to 4, the performance of SVPG does not appear to degrade. This means the policy trained using SVPG naturally adapts to a larger range of pendulums with different shapes.

The above results show that 1) SVPG can solve classical RL tasks, even with high-dimensional state space. 2) SVPG provides better robustness to the mentioned different kinds of noises and environment variations than the compared methods.

Note that the results about \textit{SVPG-shrink} and \textit{SVPG} are similar. These indicate that it is not the larger number of learnable parameters in the RWTA network that brings the above better training performance and robustness.

\subsubsection{Additional Results}
We check some more properties of SVPG.
\textbf{1) Network sparsity}. We plot the total distribution of synapse weights of 10 training trials (clipped at $ \pm 0.15 $) in the MNIST task in Figure \ref{fig:M_net_weight_dist}. Note that the number of synapse weights in SVPG-shrink and BP are close. As shown, there are more zero weights in SVPG than in BP. This indicates that SVPG tends to learn a more sparse network.
\textbf{2) Hidden layer response to input noise}. Figure \ref{fig:noise_response_distribution} shows the distribution of changes in the RWTA hidden neurons' firing probabilities $ \Delta q_h $ when adding Gaussian noise to input MNIST images (std 0.02), and its relationship with SVPG's inference iteration number. Results are gained using all 10k testing images and clipped at $ \pm 0.3 $. In early iterations, the firing probabilities are more affected by the noise as there are more of them that deviate from the no-noise firing probability; in the later iterations, they are less affected. This indicates that the recurrent design of the network should contribute to the robustness of SVPG.
\textbf{3) Computational costs}.
We summarize the times needed for inference and optimization stages in Appendix \ref{sec:computation_cost}. The rate coding version of SVPG is the fastest in optimization and is only slower than BP in inference.

\section{Discussion and Conclusion }
This paper proposes SVPG, a spiking-based variational policy gradient method with RWTA network and R-STDP. Current results reveal its potential for solving high-dimensional RL tasks and to have inherent robustness. More theoretical analysis on the cause of robustness, and more experiments in real-world control problems may improve this method and advance the research of biologically plausible methods for RL for applications in real-life scenarios. 





\subsection*{Acknowledgements}

\noindent This work was supported in part by National Natural Science Foundation of China Grant 62206151, China National Postdoctoral Program for Innovative Talents Grant BX20220167, Shuimu Tsinghua Scholar Program, Natural Science Foundation of Fujian Province of China Grant 2022J01656, and UK Royal Society Newton Advanced Fellowship Grant NAF-R1-191082. This work was undertaken in part on ARC3, part of the High Performance Computing facilities at the University of Leeds.

\bibliography{biblist}


\newpage

\appendix

\section{Proof of Theorems} \label{PT}

\subsection{Proof of Theorem 1} \label{PT1}
Recall that the mean-field inference function is
\begin{equation}
	q_{i}={\textstyle \frac{1}{Z(q_{G(i)})}\exp\{\boldsymbol{w}_{{\rm row},i}^{{\rm T}}\boldsymbol{q}+\boldsymbol{w}_{{\rm col},i}^{{\rm T}}\boldsymbol{q}+b_{i}\},}
\end{equation}
where $ Z(q_{G(i)})={\textstyle \sum_{j\in G(i)}}\exp\{\boldsymbol{w}_{{\rm row},i}^{{\rm T}}\boldsymbol{q}+\boldsymbol{w}_{{\rm col},i}^{{\rm T}}\boldsymbol{q}+b_{i}\} $,
$ i = 1,\dots,(n_h d_h +d_a) $, $ G(i) $ is the set of indices of the neurons in the same circuit as neuron $ i $, and $ \boldsymbol{w}_{{\rm row},i} $ and $ \boldsymbol{w}_{{\rm col},i} $ are respectively the $ i $-th row and column of matrix $ \boldsymbol{W} $ (in the shape of a column vector), which corresponds to the synapses connected to neuron $i$. $ b_i $ is the $ i $-th element in vector $ \boldsymbol{b} $.

For each $ w_{jk} $, There is
\begin{equation}
	\begin{aligned}\frac{\partial q_{i}}{\partial w_{jk}}= & -Z^{-2}(q_{G(i)})\frac{\partial Z(q_{G(i)})}{\partial w_{jk}}\exp\left\{ \boldsymbol{w}_{{\rm row},i}^{{\rm T}}\boldsymbol{q}+\boldsymbol{w}_{{\rm col},i}^{{\rm T}}\boldsymbol{q}+b_{i}\right\} \\
		& +Z^{-1}(q_{G(i)})\exp\left\{ \boldsymbol{w}_{{\rm row},i}^{{\rm T}}\boldsymbol{q}+\boldsymbol{w}_{{\rm col},i}^{{\rm T}}\boldsymbol{q}+b_{i}\right\} \\
		& \cdot\sum_{m=1}^{N}\left[\frac{\partial(w_{im}+w_{mi})}{\partial w_{jk}}q_{m}+(w_{im}+w_{mi})\frac{\partial q_{m}}{\partial w_{jk}}\right]\\
		= & -q_{i}Z^{-1}(q_{G(i)})\frac{\partial Z(q_{G(i)})}{\partial w_{jk}}+q_{i}\sum_{m=1}^{N}\left[\frac{\partial(w_{im}+w_{mi})}{\partial w_{jk}}q_{m}+(w_{im}+w_{mi})\frac{\partial q_{m}}{\partial w_{jk}}\right] .
	\end{aligned}
\end{equation}
For the term $ \frac{\partial Z(q_{G(i)})}{\partial w_{jk}} $, there is
\begin{equation}
	\begin{aligned}\frac{\partial Z(q_{G(i)})}{\partial w_{jk}}= & \frac{\partial}{\partial w_{jk}}\left\{ \sum_{m\in G(i)}\exp\left\{ \boldsymbol{w}_{{\rm row},m}^{{\rm T}}\boldsymbol{q}+\boldsymbol{w}_{{\rm col},m}^{{\rm T}}\boldsymbol{q}+b_{m}\right\} \right\} \\
		= & \sum_{m\in G(i)}\Big\{\exp\big[\boldsymbol{w}_{{\rm row},m}^{{\rm T}}\boldsymbol{q}+\boldsymbol{w}_{{\rm col},m}^{{\rm T}}\boldsymbol{q}+b_{m}\big]\\
		& \cdot\sum_{n=1}^{N}\big[\frac{\partial(w_{mn}+w_{nm})}{\partial w_{jk}}q_{n}+(w_{mn}+w_{nm})\frac{\partial q_{n}}{\partial w_{jk}}\big]\Big\}.
	\end{aligned}
\end{equation}
So we have
\begin{equation}\label{eq:AP_EQ_TH1_element_form}
	\begin{aligned}\frac{\partial q_{i}}{\partial w_{jk}}= & -q_{i}\sum_{m\in G(i)}\left\{ q_{m}\sum_{n=1}^{N}\big[\frac{\partial(w_{mn}+w_{nm})}{\partial w_{jk}}q_{n}+(w_{mn}+w_{nm})\frac{\partial q_{n}}{\partial w_{jk}}\big]\right\} \\
		& +q_{i}\sum_{n=1}^{N}\left[\frac{\partial(w_{in}+w_{ni})}{\partial w_{jk}}q_{n}+(w_{in}+w_{ni})\frac{\partial q_{n}}{\partial w_{jk}}\right].
	\end{aligned}
\end{equation}

Similarly, for each $ b_{j} $, there is
\begin{equation}\label{eq:AP_EQ_TH1_element_form_b}
	\frac{\partial q_{i}}{\partial b_{j}}=-q_{i}\sum_{m\in G(i)}\left\{ q_{m}\left[\frac{\partial b_{m}}{\partial b_{j}}+\sum_{n=1}^{N}(w_{mn}+w_{nm})\frac{\partial q_{n}}{\partial_{b_{j}}}\right]\right\} +q_{i}\left[\frac{\partial b_{i}}{\partial b_{j}}+\sum_{n=1}^{N}(w_{in}+w_{ni})\frac{\partial q_{n}}{\partial_{b_{j}}}\right].
\end{equation}

By respectively arranging Eq. (\ref{eq:AP_EQ_TH1_element_form}) and Eq. (\ref{eq:AP_EQ_TH1_element_form_b}) for each $ q_i $ into vectors, and combining the terms into matrices, we can get the Eq. (\ref{eq_diff_matrix}) in Theorem 1. \hfill $ \square $

\subsection{Proof of Theorem 2} \label{PT2}
The condition is the same as that in the proof of Theorem 1. The approximate differentiation of firing rate $q_{i}$ with respect to $w_{jk}$ and $b_j$ are:
\begin{equation}\label{eq_diff_approx_w}
	\begin{aligned}\frac{\partial\log(q_{i})}{\partial w_{jk}}= & \sum_{m=1}^{N}\Big[\frac{\partial(w_{im}+w_{mi})}{\partial w_{jk}}q_{m}\Big]\\
		& -\frac{1}{Z(q_{G(i)})}\sum_{m\in G(i)}\Big[\exp\{\boldsymbol{w}_{{\rm row},m}^{{\rm T}}\boldsymbol{q}+\boldsymbol{w}_{{\rm col},m}^{{\rm T}}\boldsymbol{q}+b_{m}\}\cdot\sum_{n=1}^{N}\frac{\partial(w_{mn}+w_{nm})}{\partial w_{jk}}q_{n}\Big]\\
		= & \sum_{m=1}^{N}\Big[\frac{\partial(w_{im}+w_{mi})}{\partial w_{jk}}q_{m}\Big]-\sum_{m\in G(i)}\Big[q_{m}\cdot\sum_{n=1}^{N}\frac{\partial(w_{mn}+w_{nm})}{\partial w_{jk}}q_{n}\Big],
	\end{aligned}
\end{equation}
\begin{equation}\label{eq_diff_approx_b}
	\begin{aligned}\frac{\partial\log(q_{i})}{\partial b_{j}}= & \frac{\partial b_{i}}{\partial b_{j}}-\frac{1}{Z(q_{G(i)})}\cdot\sum_{m\in G(i)}\Big[\frac{\partial b_{m}}{\partial b_{j}}\cdot\exp\{\boldsymbol{w}_{{\rm row},m}^{{\rm T}}\boldsymbol{q}+\boldsymbol{w}_{{\rm col},m}^{{\rm T}}\boldsymbol{q}+b_{m}\}\Big]\\
		= & \frac{\partial b_{i}}{\partial b_{j}}-\sum_{m\in G(i)}\Big[q_{m}\frac{\partial b_{m}}{\partial b_{j}}\Big].
	\end{aligned}
\end{equation}
Similar to the proof of Theorem 1, by respectively arranging Eq. (\ref{eq_diff_approx_w}) and Eq. (\ref{eq_diff_approx_b}) for each $ q_i $ into vectors, and combining the terms on the right hand side into matrices, we can get the Eq. (\ref{eq_diff_approx_w_matrix}) in Theorem 2. \hfill $ \square $

\newpage

\section{SVPG Algorithm Details} \label{APPENDIX_SVPG_implementation_details}
We summarize the overall working flow with our RWTA network as Algorithm \ref{alg:algorithm}. For conciseness, this algorithm does not cover the implementation of spike-based inference and STDP-based optimization. Note that these spike-related implementations can be done by simply replacing Eq.(\ref{ep_KL_derivation}) with Eq.(\ref{eq_trans_2}), and applying Eq.(\ref{eq_STDP_transformation}).

In practice, Actor-Critic (AC) based methods are more commonly used than vanilla Policy Gradient (PG). The main difference between AC and PG is that AC learns an extra value function in the place of $ r_t $ in PG. To adapt our algorithm to AC, we add a multi-layer perceptron trained with backpropagation to estimate the critic value function, and use the estimated value to replace the $ r_t $ in Eq.(\ref{eq_reinforce_gradient}).

The SVPG implemented with REINFORCE algorithm is presented in Algorithm \ref{alg:algorithm}. The equation numbers correspond to the equations in the main text.
\begin{algorithm}[H]
	\caption{SVPG with REINFORCE}
	\label{alg:algorithm}
	\textbf{Input}: Discount factor $ \gamma $. Training episode number $ N_{\rm epi} $. Inference iteration number $ N_{\rm iter} $. Learning rate $ \eta $. \\
	\textbf{Parameter}: Network shape $ n_h, d_h, d_a, d_s $. \\
	\textbf{Output}: RWTA Network parameter $ \theta $.
	
	\begin{algorithmic}[1] 
		\STATE Initialize $ \theta $.
		\FOR{Episode = 1, \dots, $ N_{\rm epi} $}
		\STATE Clear memory buffer $ \mathcal{D} $.
		\FOR{Training step t = 1, \dots, $ T $}
		\STATE Observe and encode state $ s_t $.
		\STATE Random initialize $ \boldsymbol{q}_a $ and $ \boldsymbol{q}_h $.
		\STATE Iterate Eq.(\ref{ep_KL_derivation}) until convergence.
		\COMMENT{Inference}
		\STATE Sample action $ a_t $ using $ \boldsymbol{q}_a $.
		\STATE Perform $ a_t $, observe reward $ r_t $ and new state $ s_{t+1} $.
		\STATE Store $ \langle s_t, a_t, r_t, s_{t+1}, \boldsymbol{q}, \boldsymbol{v} \rangle $ into $ \mathcal{D} $.
		\ENDFOR
		\STATE Get data from $ \mathcal{D} $.
		\STATE Calculate gradient using Eq.(\ref{eq_reinforce_gradient}), Eq.(\ref{eq_gradient}), Eq.(\ref{eq_diff_approx_w_matrix}).
		\COMMENT{Optimize}
		\STATE Update $ \theta \leftarrow \theta + \eta \nabla \theta$
		\ENDFOR
	\end{algorithmic}
\end{algorithm}

\newpage

\section{Experiment Details}
\subsection{Task Details} \label{APPENDIX_task_details}
\textit{MNIST task}. In this task, each episode includes only one time step. At the time step, the agent observes a randomly selected image from the MNIST training dataset, and selects one of the ten categories as its action. The input images are of size $ 28\times 28 $ and are in grayscale. The range of values is converted to $ [0,1] $ by dividing $ 255 $. For all the algorithms compared, the input images are stretched to vectors (with length $ 784 $). A reward of $ +1 $ indicates a correct selection, $ -1 $ the opposite. The maximum length of training is set to 50k steps (in each step a randomly selected 100-size batch is used for training) so that in practice all the methods converge. During training, a checkpoint of the network parameters is saved every 100 training steps. After training, the checkpoint with the best accuracy in the testing set (without noise) is reloaded to do the testing. The MNIST testing dataset is used in testing.

\textit{GymIP task}. Each episode has a maximum of $ 200 $ time steps, with a reward of $ +1 $ for each step. The episodes end early if the pendulum (pole) falls. The observation is a $ 4 $-dimensional vector with no predefined ranges. To normalize the observations to the range of $ [0,1] $, we use a random policy to sample from the environment, and use the samples' range to determine a linear mapping to the range of $ [0,1] $. In our experiments, the sampled ranges are $ [-0.4, 0.4] $, $ [-0.2, 0.2] $, $ [-1.7, 1.7] $, $ [-1.25, 1.25] $. The action space consists of 5 discrete actions, evenly extracted from the range $ [-3, 3] $. The maximum length of training is set to 20k episodes. During training, a checkpoint of the network parameters is saved every 20 episodes. After training, the checkpoint with the best performance is reloaded to do the testing.

\subsection{Implementation Details} \label{sec:Appendix_alg_details}
\textbf{Network Sizes.} 1) For the MNIST task, SVPG uses an RWTA network with 784 input neurons, 20 hidden WTA circuits each with 10 neurons, and 10 output neurons; SVPG-shrink uses a smaller RWTA network where the number of hidden WTA circuits is changed to 17 so that the total number of learnable parameters is close to other methods; BP, BPTT, ANN2SNN, and EP use layered networks with 784 input neurons, 1 hidden layer with 200 neurons, and 10 output neurons. 2) For the GymIP task, SVPG uses an RWTA network with 4 input neurons, 8 hidden WTA circuits each with 8 neurons, and 5 output neurons; SVPG-shrink uses a smaller RWTA network where the number of hidden WTA circuits is changed to 3. BP, BPTT, and ANN2SNN use layered networks with 4 input neurons, 1 hidden layer with 64 hidden neurons, and 5 output neurons. 3) For the GymIP task there is a critic network used in all the methods. This critic network is layered, with 4 input neurons, 2 hidden layers each with 64 neurons, and 5 output neurons.

\textbf{Optimizer.} For the compared methods, i.e., BP, BPTT, and EP, we select the stochastic gradient descent (SGD) with zero momentum for the MNIST task, and select the RMSprop optimizer for the GymIP task. For the SGD optimizer, we incompletely tried learning rates ranging from 0.001 to 0.3, and found 0.1 to be a good balance between training speed and stability. For the RMSprop optimizer, we use a learning rate of 0.001. As for SVPG, we use a learning rate of 0.1 in the MNIST task and 0.001 in the GymIP task.

\textbf{RL hyper-parameters.} The GymIP task involves sequential decisions which may have long-term effects on rewards. We select the discount rate $ \gamma $ to be 0.999 (close to 1) so that the discounted sum of rewards reflects the length of the episodes, maximizing which is the task objective. To stabilize training, we use a replay buffer with a size of 100. To encourage the agent to explore more actions, we add an intrinsic exploration reward to the environment reward; the reward is calculated as the entropy of the agent's action distribution; the ratio of the environment reward and the intrinsic reward is 2:1.

\newpage

\section{Additional Experiment Results}
\subsection{MNIST Input Noise Illustration} \label{sec:MNIST_noise}

\begin{figure}[H]
	\centering
	\begin{minipage}[b]{0.6\linewidth}
		\vskip -3mm
		{\small\quad\ 0\ \ \  \hfill 0.2 \hfill 0.4 \hfill 0.6 \hfill 0.8 \hfill 1.0 \ \ \ } \\
		\includegraphics{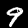}
		\hfill
		\includegraphics{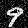}
		\hfill
		\includegraphics{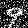}
		\hfill
		\includegraphics{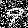}
		\hfill
		\includegraphics{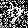}
		\hfill
		\includegraphics{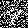}
		
		\vskip -3mm
	\end{minipage}
	\caption{Some input images with different strengths of Gaussian noises in MNIST task. Standard deviation noted above images.}
	\label{fig:MNIST_input_noise}
\end{figure}

\subsection{Additional Comparison of Three SVPG Implementations} \label{sec:Additional_comparison_implementation}
\begin{figure}[H]
	\centering
	\subfigure[Input Gaussian]{
		\centering
		\includegraphics[height=23mm]{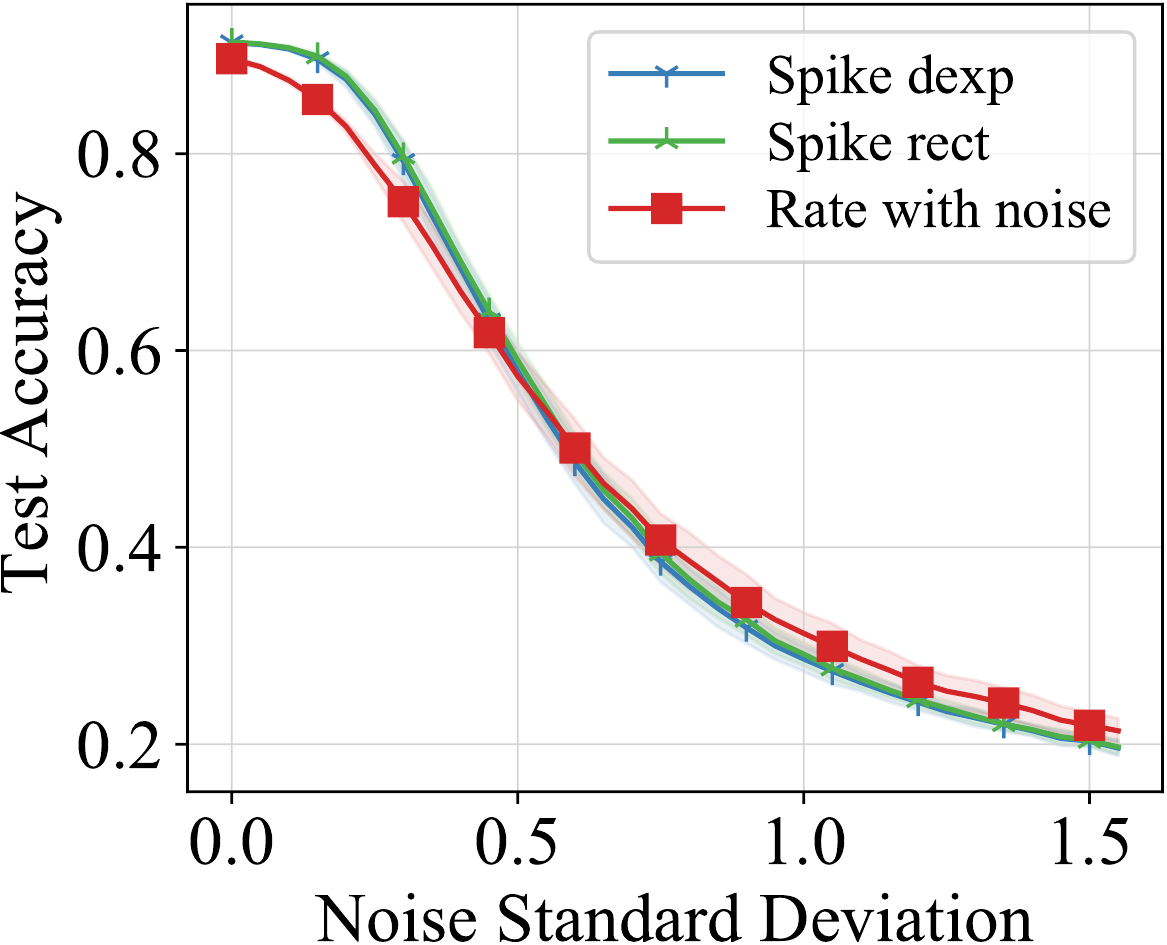}
	}
	\subfigure[Input Gaussian\&Salt]{
		\centering
		\includegraphics[height=23mm]{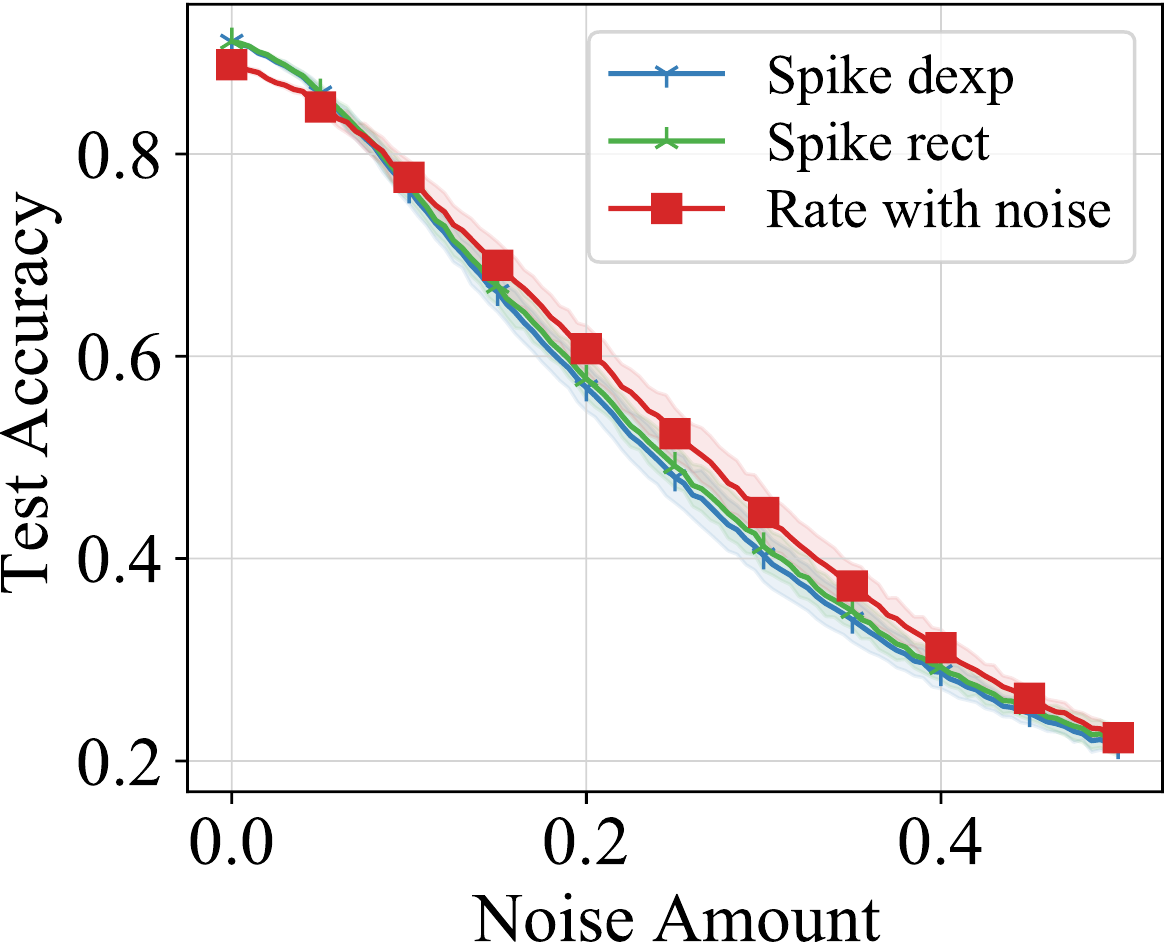}
	}
	\subfigure[Input Salt\&Pepper]{
		\centering
		\includegraphics[height=23mm]{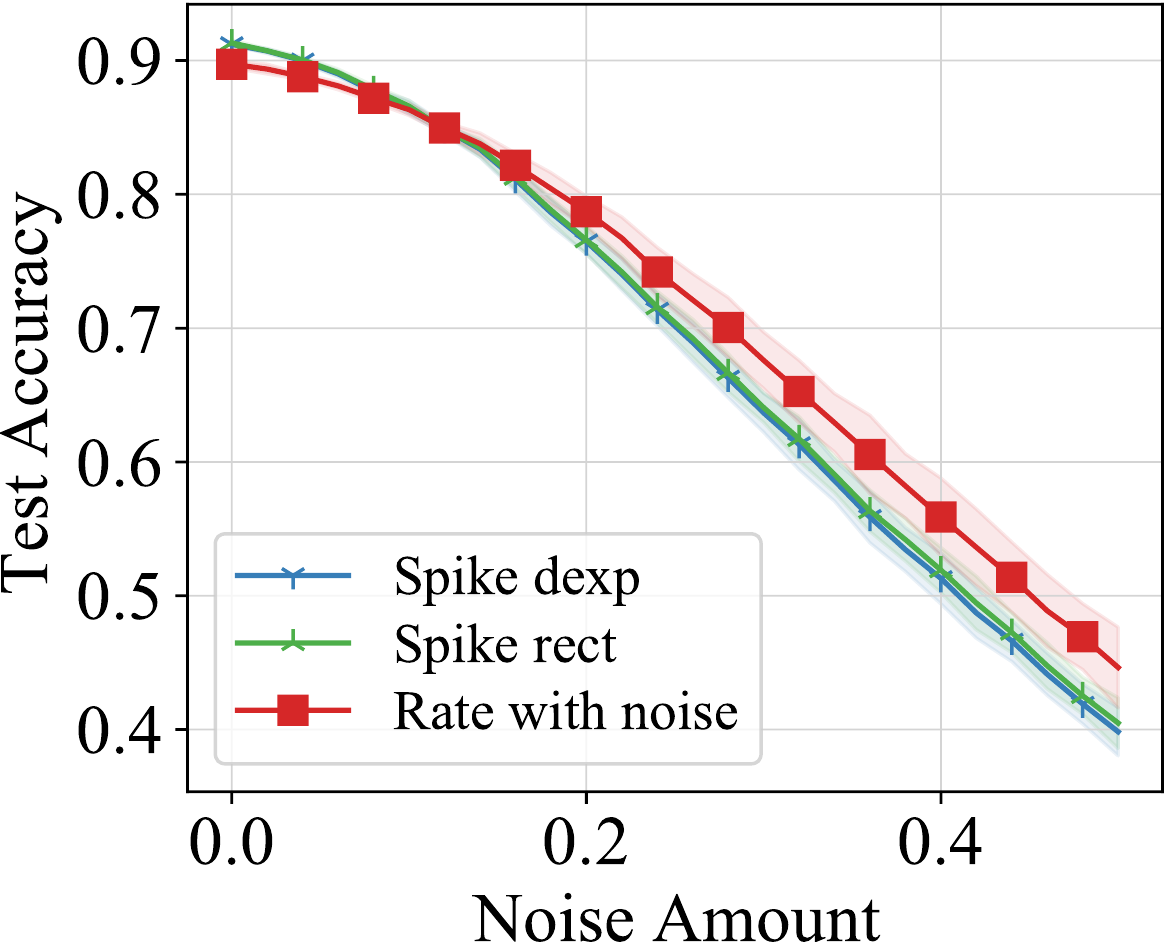}
	}
	\subfigure[Network Uniform]{
		\centering
		\includegraphics[height=23mm]{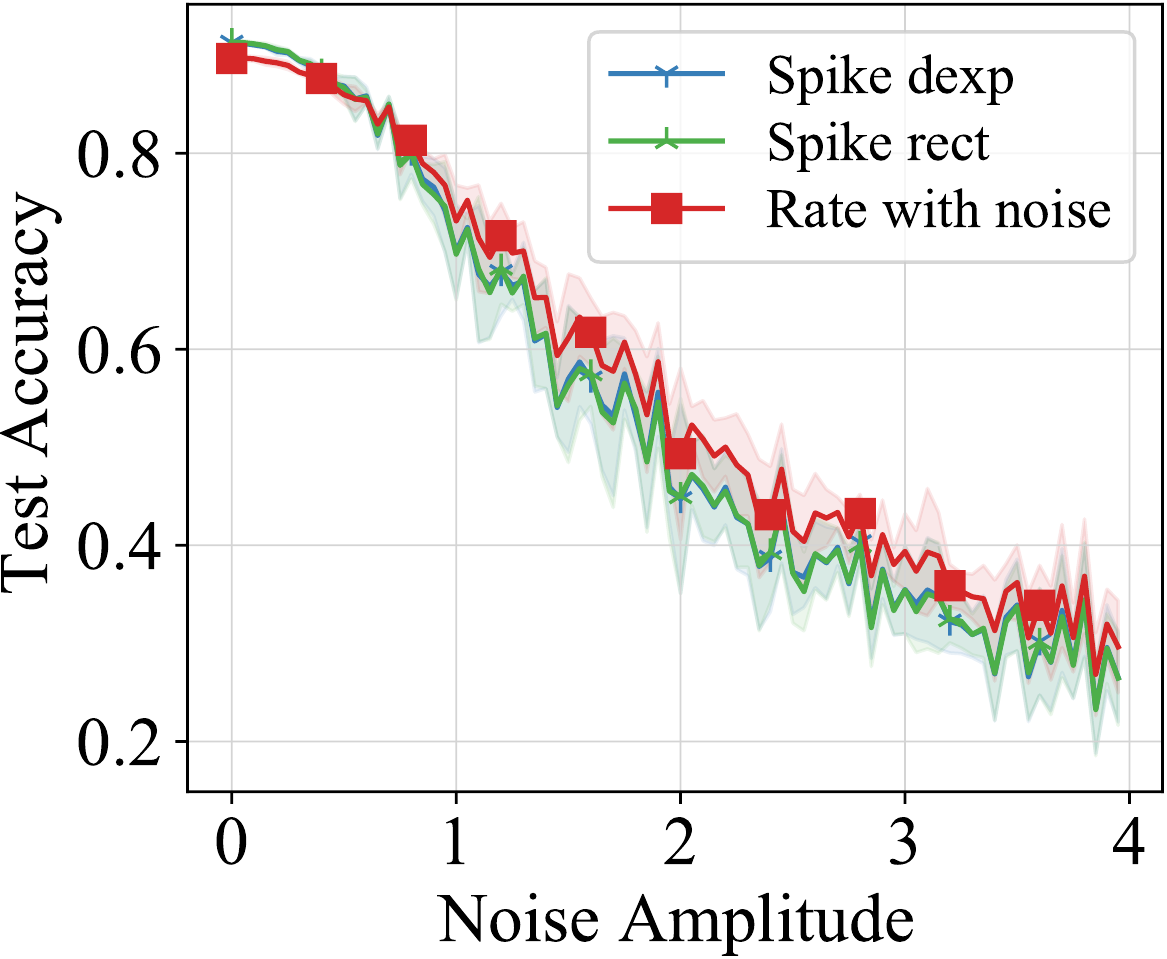}
	}
	\caption{Additional comparison of three implementations of SVPG.}
	\label{fig:SVPG_compare_others}
\end{figure}

In the main text, we present the results regarding input salt noise and network Gaussian noise. Here we present the results with other noises, as shown in Figure \ref{fig:SVPG_compare_others}. These results further support the analysis in the main text.

\subsection{MNIST Additional Results}
\begin{figure}[H]
	\centering
	\subfigure[Gaussian]{
		\centering
		\includegraphics[height=23mm]{draw/exp_figures/I_gaussian}
	}
	\hfill
	\subfigure[Salt]{
		\centering
		\includegraphics[height=23mm]{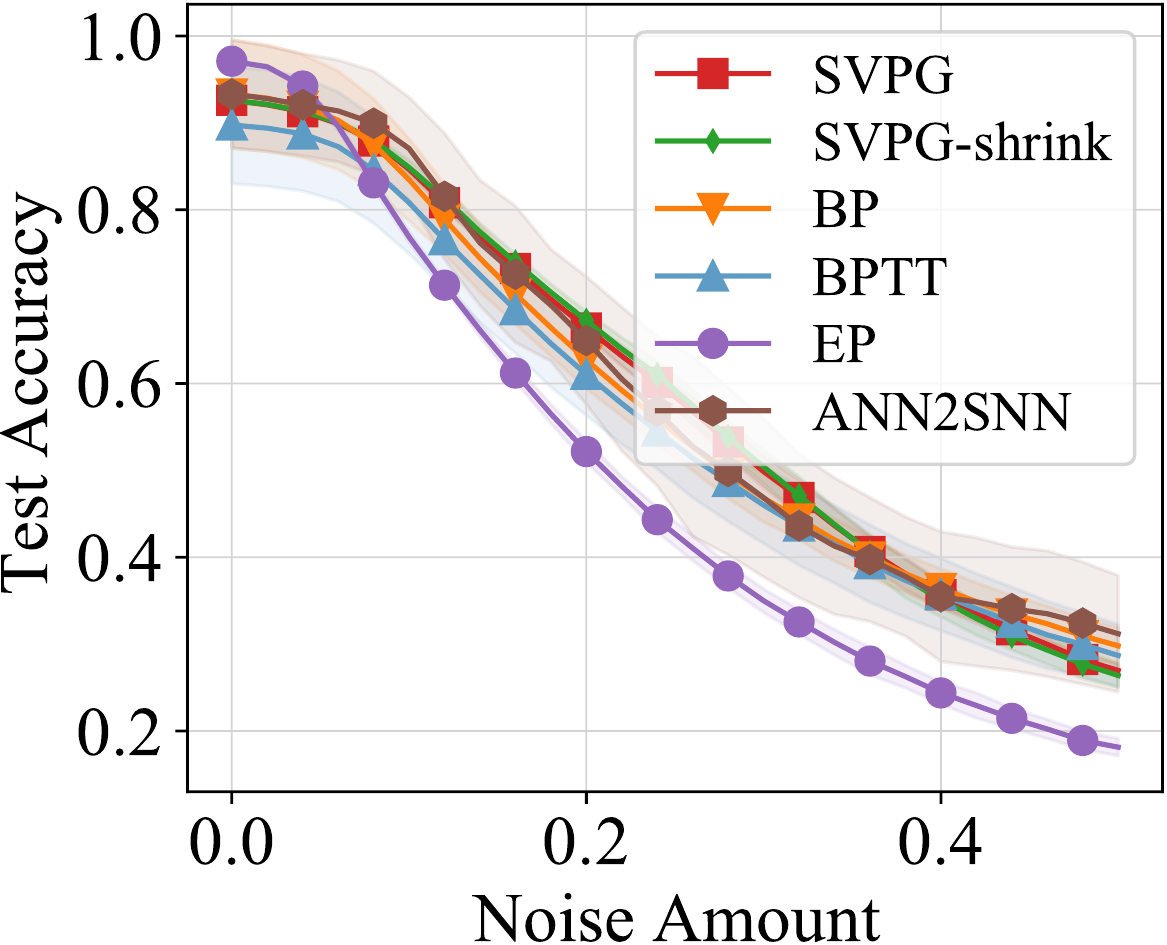}
	}
	\hfill
	\subfigure[Salt\&Pepper]{
		\centering
		\includegraphics[height=23mm]{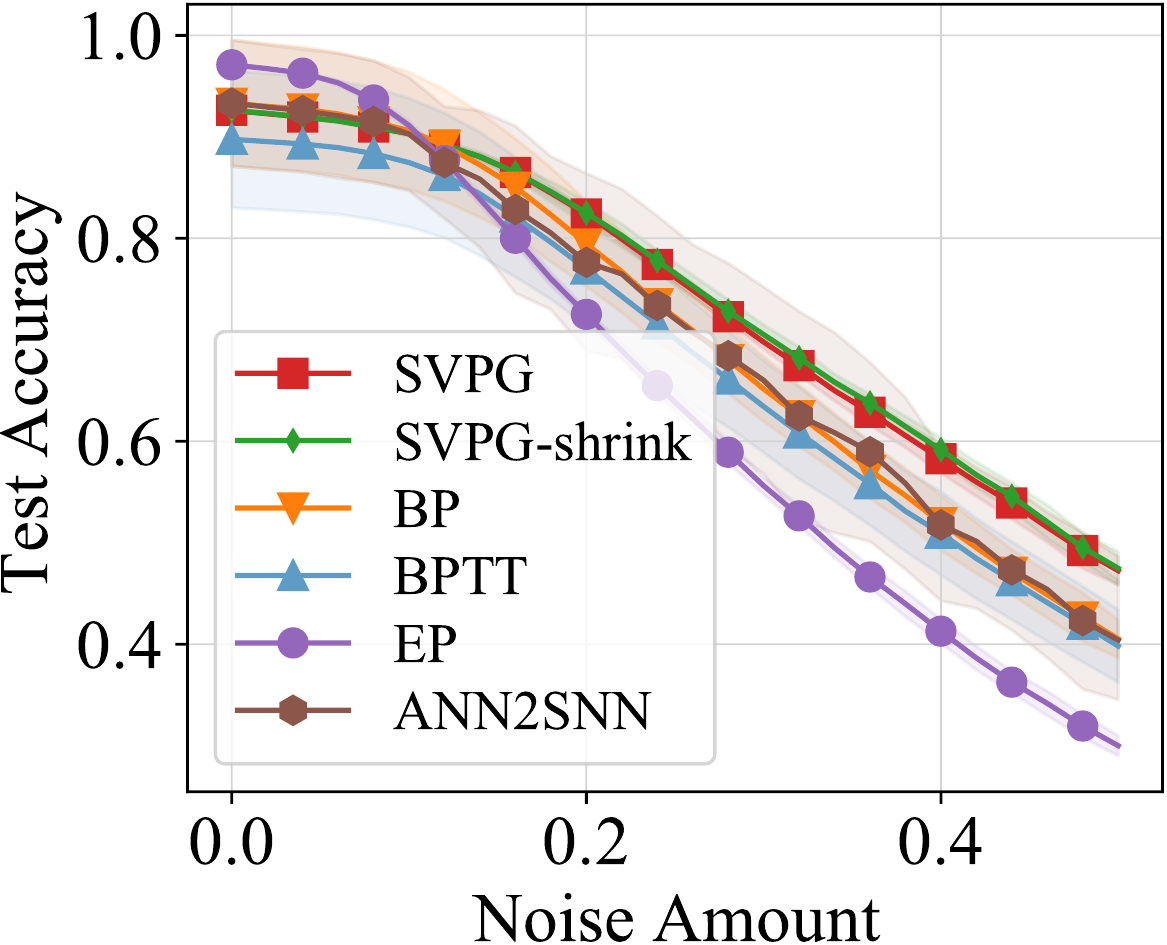}
	}
	\hfill
	\subfigure[Gaussian\&Salt]{
		\centering
		\includegraphics[height=23mm]{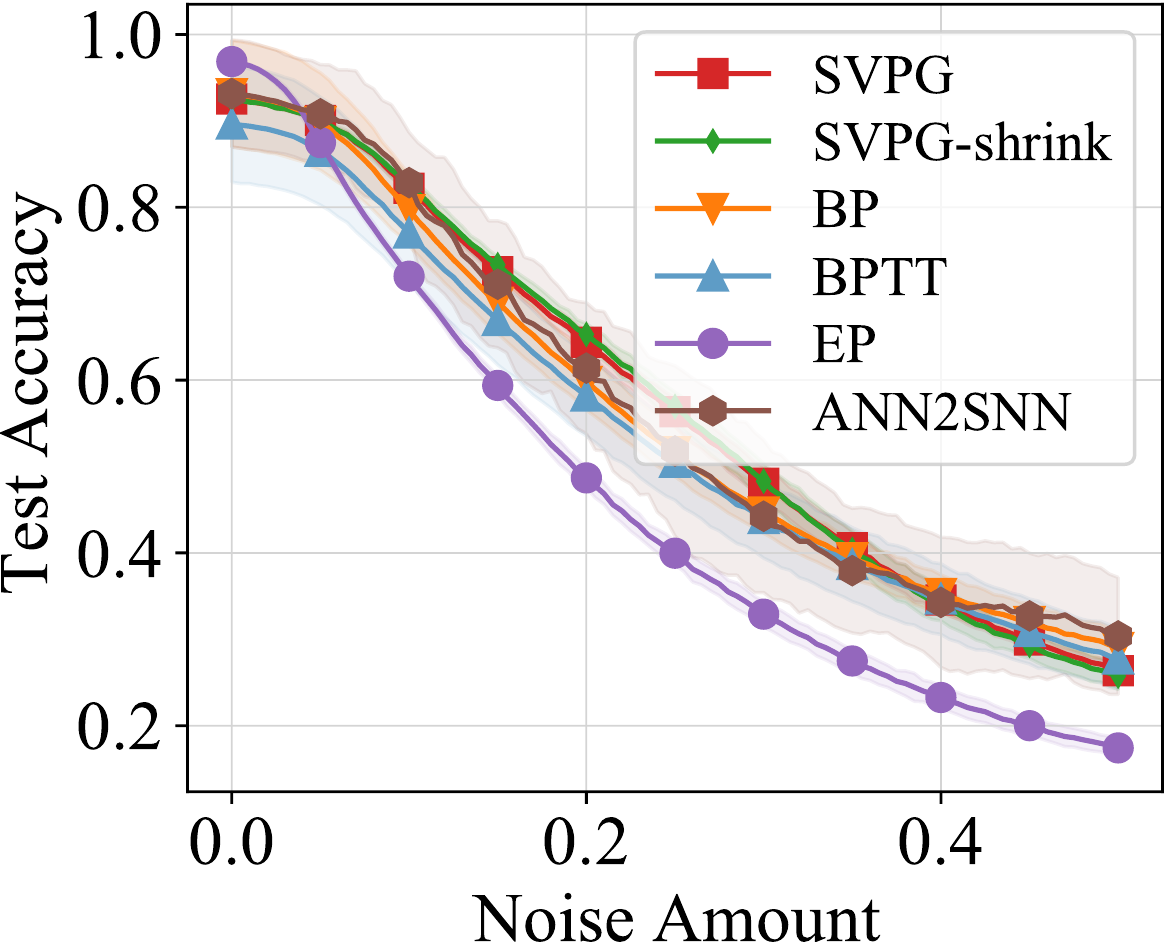}
	}
	\caption{MNIST -- Input noises.}
	\label{fig:M_I_noise}
\end{figure}

\begin{figure}[H]
	\centering
	\subfigure[Gaussian]{
		\centering
		\includegraphics[height=23mm]{draw/exp_figures/W_gaussian}
	}
	\subfigure[Uniform]{
		\centering
		\includegraphics[height=23mm]{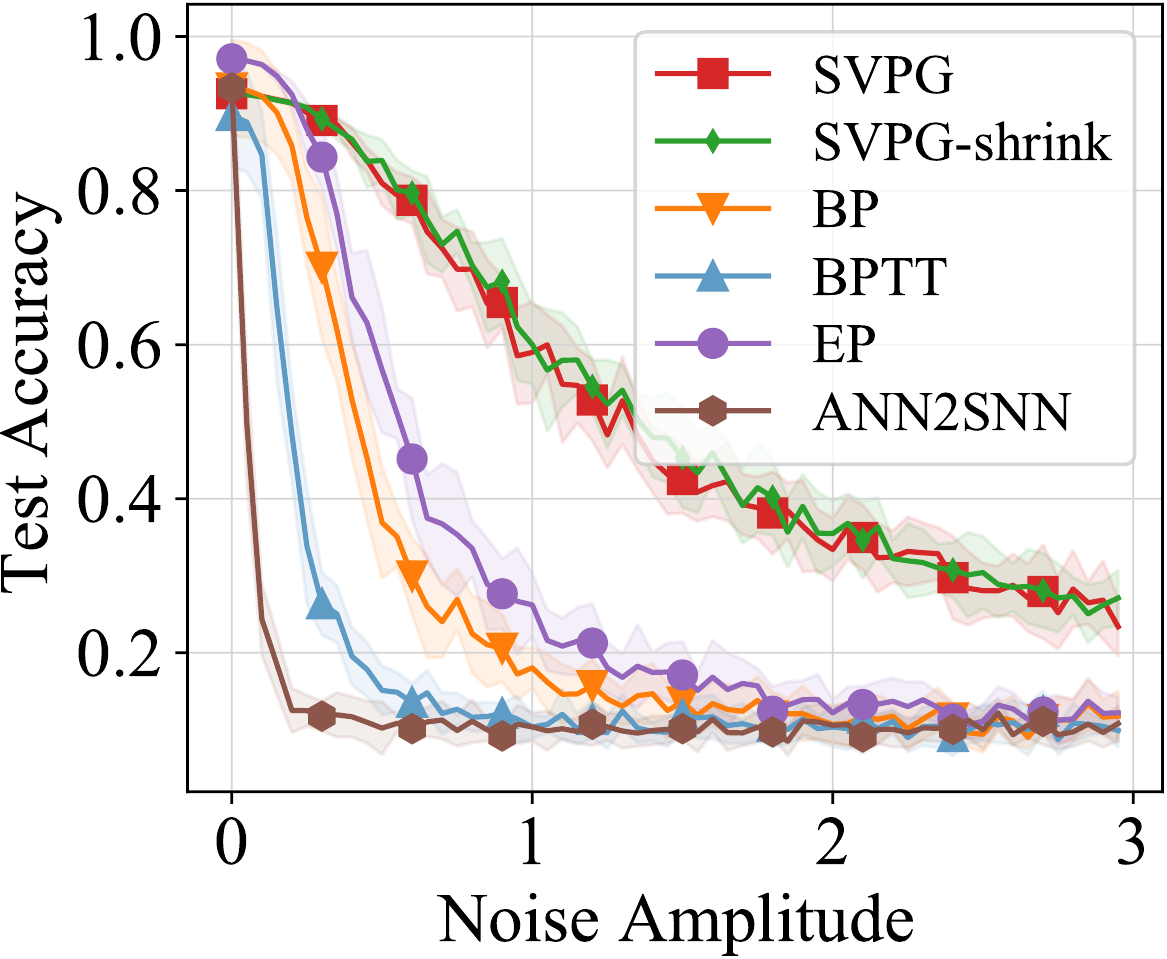}
	}
	\caption{MNIST -- Network noises.}
	\label{fig:M_W_noise}
\end{figure}

\subsection{GymIP Additional Results}
Here are the complete results of all the variations tested in the GymIP task. Note that the ``Union'' variation means the length and thickness of the pendulum change together with a fixed ratio (length:thickness=16:1).

\begin{figure}[H]
	\centering
	\begin{minipage}{0.49\linewidth}
		\begin{figure}[H]
			\centering
			\subfigure[Gaussian]{
				\centering
				\includegraphics[height=23mm]{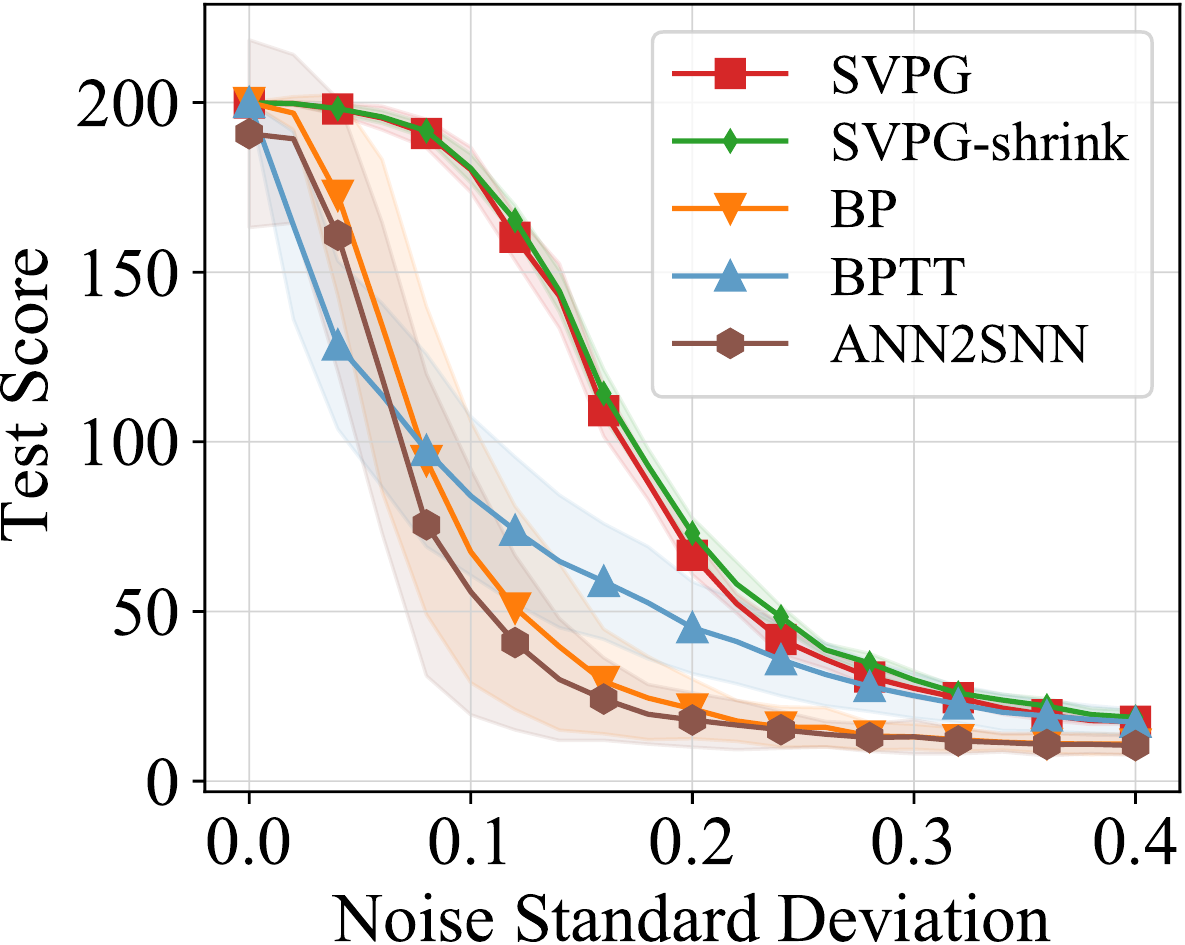}
			}
			\subfigure[Uniform]{
				\centering
				\includegraphics[height=23mm]{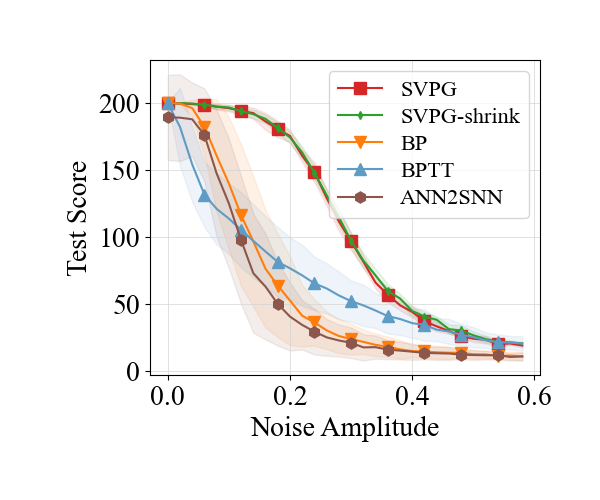}
			}
			\caption{GymIP -- Input noises.}
			\label{fig:G_I_noise}
		\end{figure}
	\end{minipage}
\hfill
	\begin{minipage}{0.49\linewidth}
		\begin{figure}[H]
			\centering
			\subfigure[Gaussian]{
				\centering
				\includegraphics[height=23mm]{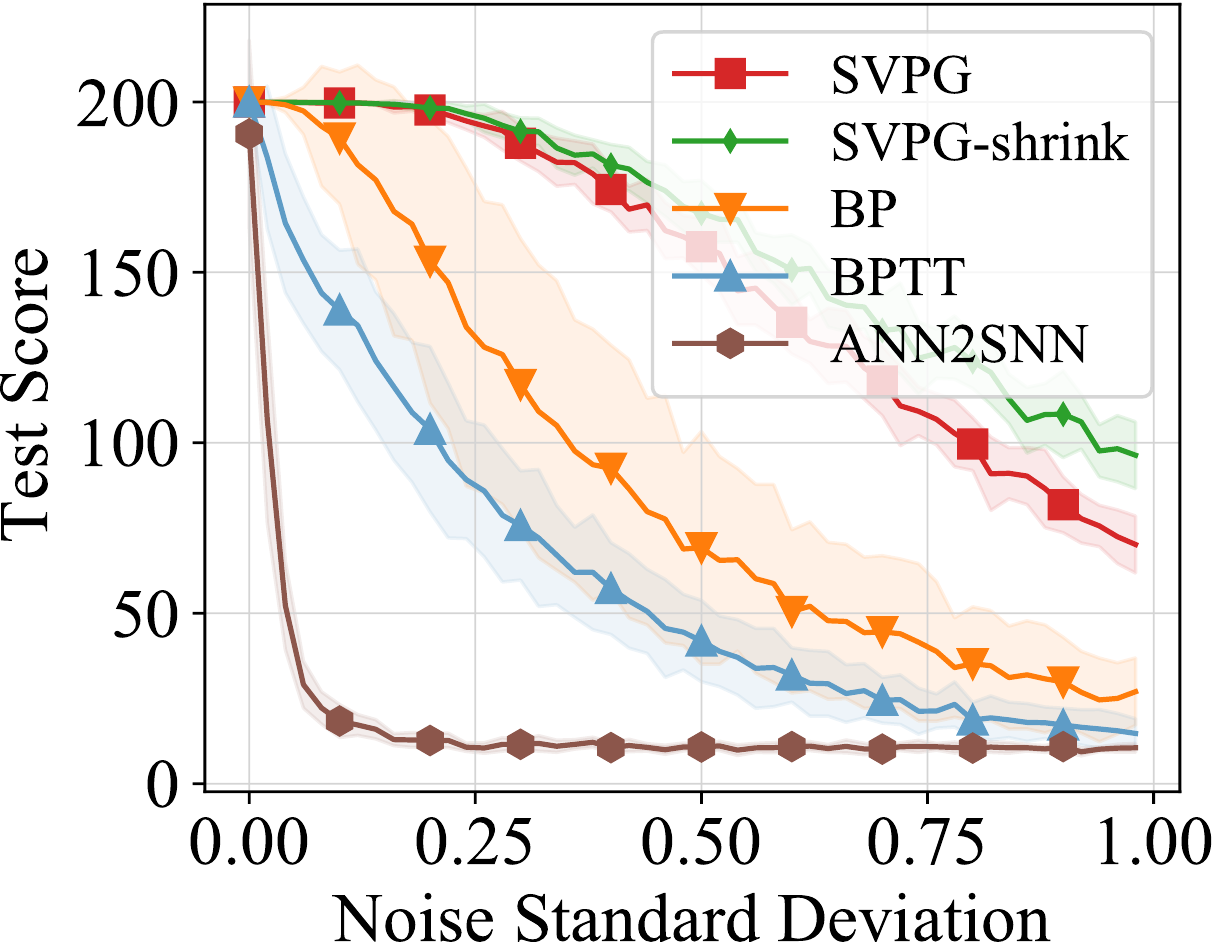}
			}
			\subfigure[Uniform]{
				\centering
				\includegraphics[height=23mm]{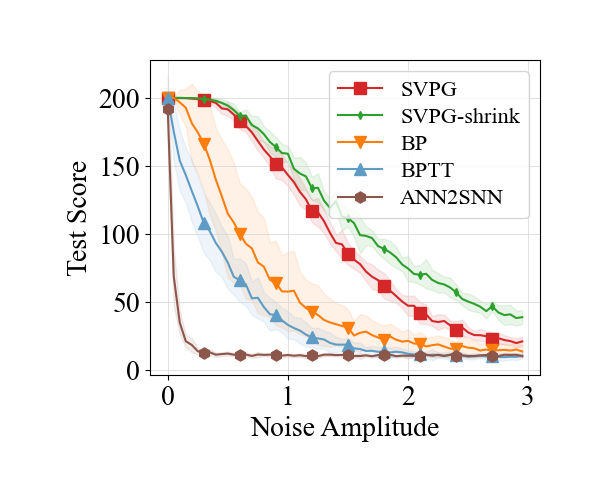}
			}
			\caption{GymIP -- Network noises.}
			\label{fig:G_W_noise}
		\end{figure}
	\end{minipage}
\end{figure}

\begin{figure}[H]
	\centering
	\subfigure[Length]{
		\centering
		\includegraphics[height=23mm]{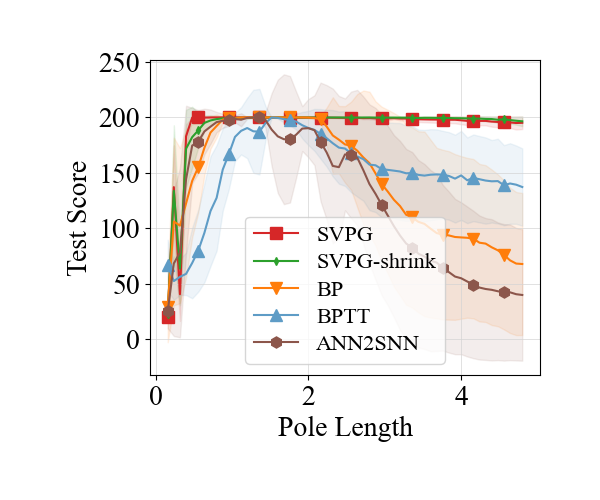}
		\label{fig:robust_env_GIP_1}
	}
	\subfigure[Thickness]{
		\centering
		\includegraphics[height=23mm]{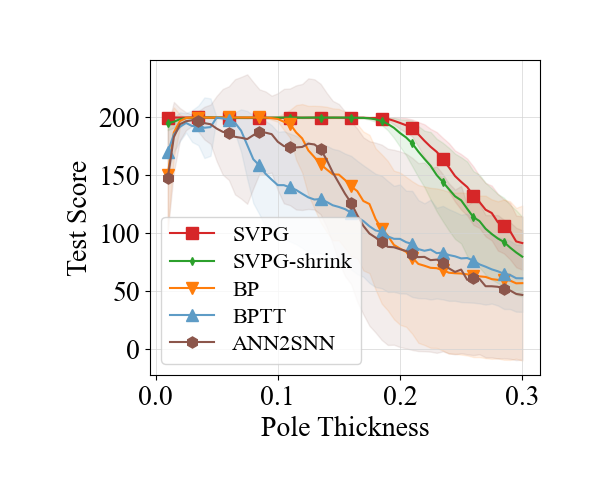}
		\label{fig:robust_env_GIP_2}
	}
	\subfigure[Union]{
		\centering
		\includegraphics[height=23mm]{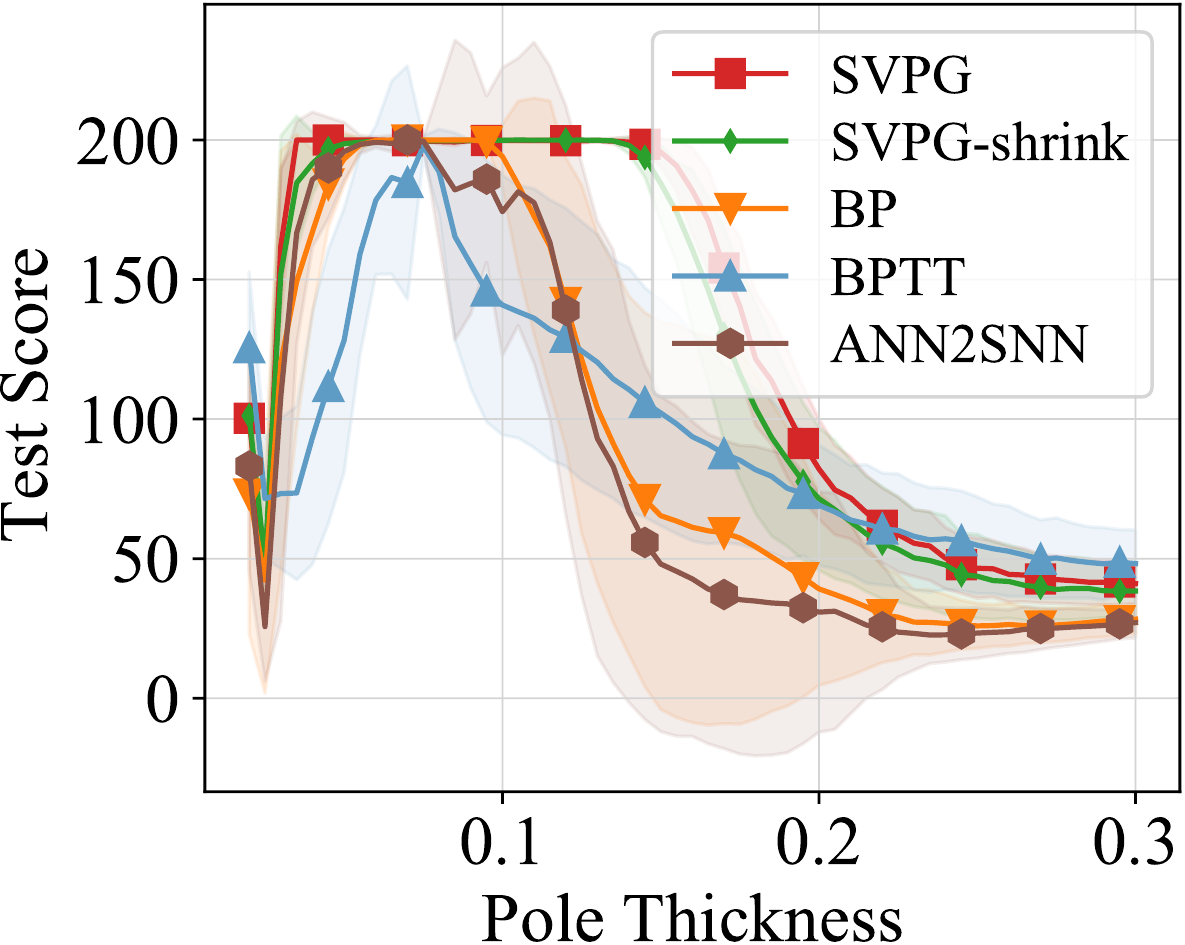}
		\label{fig:robust_env_GIP_3}
	}
	\caption{GymIP -- Pendulum variations.}
	\label{fig:robust_env_GIP}
\end{figure}

\subsection{Computation Costs} \label{sec:computation_cost}
The computation costs of inference and optimization of different methods in the MNIST task are shown in Table \ref{table:computation_costs}. The values presented are the averaged times of 10 steps. The results are gained using one NVIDIA RTX 3080 GPU.
1) For the inference period, SVPG consumes much more time than BP and EP. This is because SVPG needs to simulate a spike train during each inference step. The rate coding version of SVPG alleviates this problem and achieves a computational efficiency close to that of BP and EP. 2) For the optimization period, SVPG is the most efficient. This is because SVPG updates the parameters using only local learning rules, while other methods need backpropagation (BP and BPTT) or iterations (EP).

\begin{table}[H]
	\centering
	\begin{tabular}{cccccc}
		\toprule 
		Time (ms)    & SVPG-rate & SVPG-spike & BP  & BPTT & EP \tabularnewline
		\midrule
		Inference    & 2.1       & 373.1      & 0.2 & 52.5 & 4.3 \tabularnewline
		\midrule 
		Optimization & 0.2       & 0.5        & 1.6 & 60.1 & 61.1 \tabularnewline
		\bottomrule
	\end{tabular}
	\vskip 3mm
	\caption{Computation costs.}
	\label{table:computation_costs}
\end{table}

%


\end{document}